\documentclass[lettersize,journal]{IEEEtran}
\usepackage{amsmath,amsfonts}
\usepackage{algorithm}
\usepackage{enumerate}

\usepackage{array}
\usepackage[table]{xcolor}
\usepackage{caption}
\usepackage{textcomp}
\usepackage{stfloats}
\usepackage{url}
\usepackage{verbatim}
\usepackage{graphicx}
\usepackage{hyperref}
\usepackage{tikz}
\usepackage{subcaption}
\usepackage{multirow}
\usepackage{svg}
\usepackage{diagbox}
\usepackage{comment}

\usepackage[ font=footnotesize]{caption}

\usepackage[textwidth=40]{todonotes}

\usepackage{cite}
\newtheorem{definition}{\textbf{\textit{Definition}}}

\newtheorem{rmk}{\textbf{\textit{Remark}}}

\usepackage{algpseudocode}
\usepackage{dirtytalk}

\algnewcommand{\algorithmicforeach}{\textbf{for each}}
\algdef{S}[FOR]{ForEach}[1]{\algorithmicforeach\ #1\ \algorithmicdo}

\definecolor{tableborder}{RGB}{0, 0, 255} % Define the color (RGB values for blue)

\begin{document}

\title{RACP: Risk-Aware Contingency Planning with Multi-Modal Predictions}

\author{ Khaled A. Mustafa, Daniel Jarne Ornia, Jens Kober, and Javier Alonso-Mora
\thanks{ The authors are with the Dept. of Cognitive Robotics, TU Delft, 2628 CD Delft, The Netherlands, {\tt\small $\{$k.a.mustafa; d.jarneornia; j.kober; j.alonsomora$\}$@tudelft.nl}.
}
\thanks{This research is supported by funding from the Dutch Research Council NWO-NWA, within the “Acting under Uncertainty” (ACT) project (Grant No. NWA.1292.19.298).}
}

\markboth{IEEE Transactions on Intelligent Vehicles}%
{Shell \MakeLowercase{\textit{et al.}}: A Sample Article Using IEEEtran.cls for IEEE Journals}


\maketitle

\begin{abstract}
For an autonomous vehicle to operate reliably within real-world traffic scenarios, it is imperative to assess the repercussions of its prospective actions by anticipating the uncertain intentions exhibited by other participants in the traffic environment. Driven by the pronounced multi-modal nature of human driving behavior, this paper presents an approach that leverages Bayesian beliefs over the distribution of potential policies of other road users to construct a novel risk-aware probabilistic motion planning framework. In particular, we propose a novel contingency planner that outputs long-term contingent plans conditioned on multiple possible intents for other actors in the traffic scene. The Bayesian belief is incorporated into the optimization cost function to influence the behavior of the short-term plan based on the likelihood of other agents' policies. Furthermore, a probabilistic risk metric is employed to fine-tune the balance between efficiency and robustness. Through a series of closed-loop safety-critical simulated traffic scenarios shared with human-driven vehicles, we demonstrate the practical efficacy of our proposed approach that can handle multi-vehicle scenarios.
\end{abstract}


\section{Introduction}
\IEEEPARstart{S}{afe} motion planning is a prominent feature in the self-driving stack. In urban scenarios, the ego-agent needs to understand and infer the intended motion of other road users in the scene in order to move safely and efficiently. However, predicting the behavior of road users poses great challenges since they exhibit non-deterministic and multi-modal behaviors. Moreover, their intentions cannot be explicitly communicated to the ego-agent. For instance, in a non-signalized intersection, it is hard to anticipate whether a human driver will drive straight or turn right, and it is crucial to encode this uncertainty into the planning formulation. This gives rise to stochastic prediction models that provide probabilistic information over all possible intentions the human driver can exhibit \cite{Salzmann, Rhinehart, Casas, Y Chen}. By leveraging this probabilistic information, the ego-agent's motion planner needs to generate safe trajectories yet not overly conservative in the presence of other road users.

An important aspect of planning under uncertainty is to guarantee the existence of collision-free trajectories despite the stochastic motion of the surrounding obstacles. One class of methods that deals with the uncertain behavior of dynamic agents is \textit{robust optimization} \cite{bental} which provides safety guarantees by rigorously accounting for bounded sets of uncertainties. That is, the probability density function of the uncertainty is non-zero over a bounded domain of the ego agent’s workspace and is zero elsewhere. However, since robust optimization accounts for all possible realizations of the uncertainty, its behavior is excessively conservative and may result in infeasible solutions in crowded environments, a well-studied issue in robot navigation literature known as the ``frozen robot'' problem \cite{Trautman}. In contrast, \textit{stochastic optimization} \cite{Mesbah} uses \textit{chance constraints} \cite{Zhu, Wange, de groot} to loosen hard constraints and bound the probability of violating safety constraints to be within a desired confidence level $\delta$. This, in turn, results in less-conservative behavior compared to robust optimization approaches. However, in multi-modal traffic scenarios, this can still result in conservative behavior since a single trajectory is sought to minimize the optimization's cost function along the entire planning horizon. This gives rise to contingency planning frameworks \cite{Hardy, Bajcsy, Chen}, that explicitly plan a set of conditional actions that depend on the stochastic outcome of a prediction model.

\begin{figure}[!t]
\centering
\includegraphics[width=2.8in]{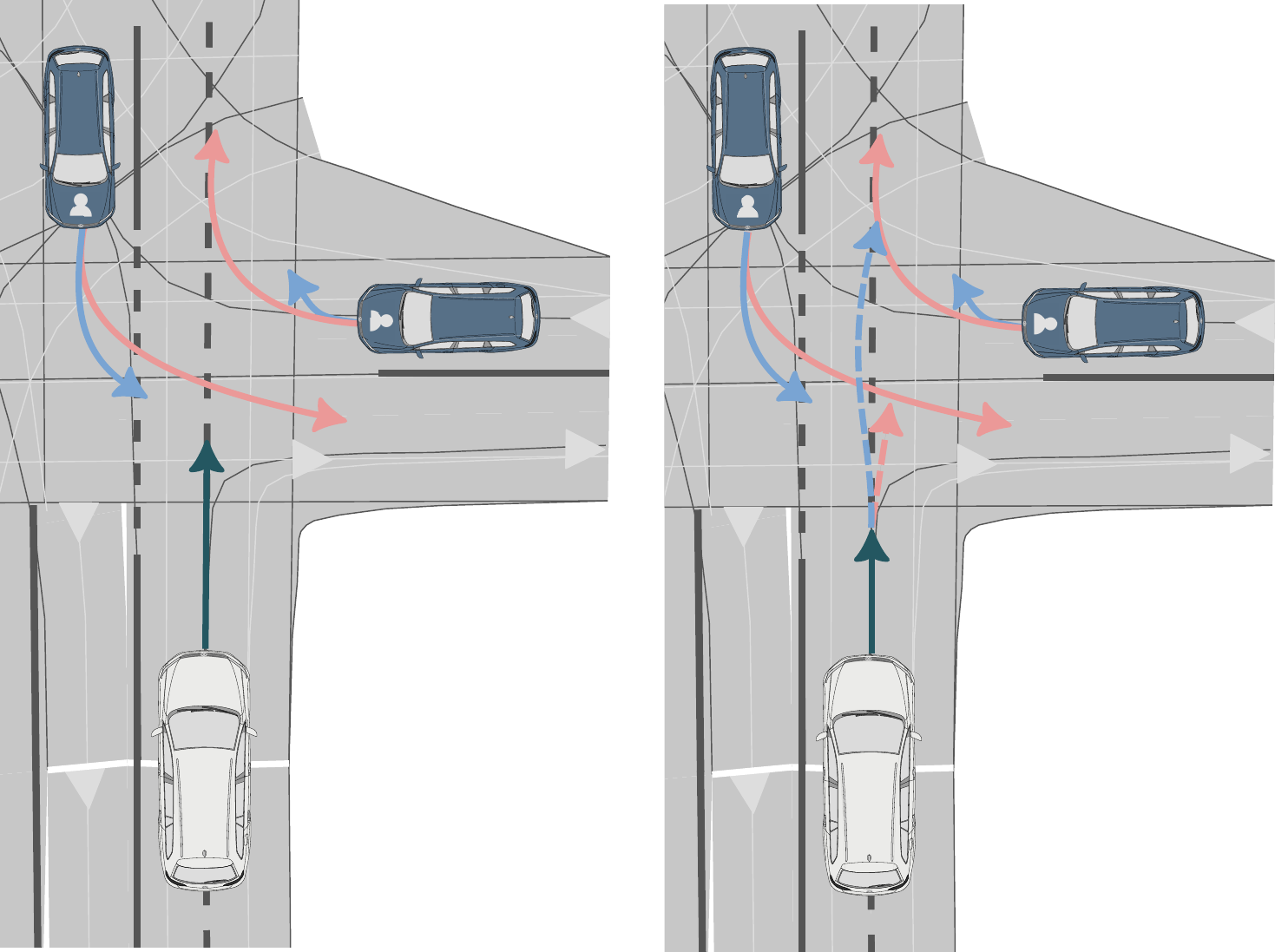}
\caption{Two of the possible future intents of the human-driven vehicle are shown in red and blue. On the left, the ego-vehicle seeks a single plan which is safe with respect to both intents. On the right, a short-term trajectory is planned that branches into two contingent plans for each human's intents. The illustrative example is inspired by \cite{Urtasun}.}
\label{fig_1}
\end{figure}

As an illustrative example, to highlight the difference between single policy planning and contingency planning, consider the three-way intersection scenario depicted in Fig.~\ref{fig_1}. In this scenario, we consider only two possible intents a human driver can express. The one depicted in blue shows that the human driver yields to the autonomous vehicle, whereas the red one indicates that the autonomous vehicle brakes since the human driver takes an aggressive left turn. A traditional planner, in that situation, seeks a single plan that is safe with respect to both intents resulting in a braking trajectory. In contrast, since only one of the predicted intentions will happen in the future, the contingency planner plans a short-term trajectory that ensures safety for both potential outcomes. Subsequently, it diverges into two specific plans, each tailored to address a distinct future intention resulting in less-conservative plans. 

\section{Related Work And Contribution}
\subsection{Related Work}
A primary objective of the
motion planner is to generate non-conservative, yet safe trajectories,
for the ego vehicle to execute. In some of the proposed methods in the literature, the planner optimizes for the worst-case scenario, of how the motion of other road users will propagate into the future, regardless of its likelihood \cite{Xu}. Despite being safe, this causes the ego vehicle to behave defensively and overreact to low-probability dangers far into the future, e.g., the ego vehicle brakes prematurely to react to an unlikely future which would be uncomfortable and socially confusing for other road
users. Human behavior, in these situations, would not be
either overly conservative or completely ignorant of such rare
scenarios. An alternative way is to generate a single plan, that safeguards against all possible intents along the entire planning horizon which results in inefficient and conservative plans with compromised performance \cite{Vasudevan, Murray}. \textcolor{black}{As a consequence, uncertain scenarios require autonomous vehicles to find a reasonable trade-off between safety and efficiency. This gives rise to the problem of \textit{planning with multi-modal predictions}.}
\subsubsection{\textcolor{black}{Planning with multi-modal predictions}}
Fail-safe motion planning was introduced in \cite{Magdici, Pek Fail-safe, Pek} where a single trajectory is generated by considering the most probable trajectories of the other agents. Then, the safety of the proposed method is guaranteed by ensuring the existence of an emergency maneuver at each time step that accounts for every possible trajectory of the other agents. Although the fail-safe trajectory may ensure safety on a finite horizon plan, recursive feasibility is not guaranteed. Moreover, it is difficult to estimate whether the ego-vehicle is close to a collision.

Legibility-based models are widely used in the literature to alleviate the conservatism of planning under uncertainty \cite{Srinivasa, Brudigam, Van der Loos}. A motion is considered legible if it allows an observer to confidently infer the correct agent's intent after observing a snippet of its trajectory \cite{Dragan}, and the legibility of the motion depends on the required time until an observer can infer an agent's intent. In these approaches, a probabilistic model is used to infer the probability of a certain goal $G \in \mathcal{G}$ from an incomplete initial trajectory $\zeta_{S \rightarrow Q}$, $P(G|\zeta_{S \rightarrow Q})$, and the agent's inferred goal is modeled as the most likely one $\arg \max_{G\in \mathcal{G}}P(G|\zeta_{S \rightarrow Q})$. This may, however, result in an over-confident plan leading to a collision since the probability that the agent moves towards a different goal is entirely ignored. 

Branch Model Predictive Control (B-MPC), on the other hand, is utilized in \cite{Alsterda, Batkovic, Chen, Oliveira} to tackle the multi-modality arising from human-driving decision-making, where the behavior of the surrounding agents is simplified with a finite set of policies derived from a prediction model. A probabilistic scenario tree is then constructed from this finite set where each branch in the tree has an associated policy. On top of the scenario tree, a trajectory tree is built that shares the same topology as the scenario tree where the objective is to minimize the expected cost over all the branches. Yet, these approaches suffer from the \textit{curse of dimensionality} since the tree structure grows exponentially with the prediction horizon and the number of agents, making it only feasible for short planning horizons \cite{Fors}.

\textcolor{black}{Another line of work for planning under uncertainty is by formulating the planning problem as a partially observable Markov decision process (POMDP) by constructing a belief tree based on a discrete set of obstacle vehicle's intentions \cite{Liu, Hubmann}. Solving such problems, however, becomes computationally intractable when the problem size scales. To address this problem,} multi-policy decision-making \cite{MPDM, Galceran, Zhang-pomdp} decomposes the belief tree into a limited number of closed-loop policies by leveraging semantic information. However, these approaches solve for the best ego-policy over all possible future realizations resulting in overly-conservative plans that do not exploit the multi-modality in the obstacle-vehicle behavior. To tackle this problem, inspired by branch-MPC, TPP \cite{TPP} proposes an approach that converts the continuous space motion planning problem into a tractable problem by converting both the trajectory tree and scenario tree into a finite-horizon MDP. The optimal policy is then determined via dynamic programming over the constructed MDP. Despite its scalability to multiple vehicles, TPP outputs a single optimal policy over all scenarios causing unavoidable loss of multi-modality information.

This issue can be tackled by separating the planning problem into short-term and long-term responses \cite{Hardy, Zhan, Bajcsy, Urtasun}. This is realized by generating a short-term trajectory, for $t<t_b$, where $t_b$ indicates a branching time, which guarantees safety with respect to all possible future intents that the other agents can exhibit. After $t_b$ duration, it is assumed that the ego vehicle will be able to determine the intent of the human driver, and thus it is sufficient to safeguard against the most likely intent. Thus, the contingent plan can ensure safety only when the ego vehicle acquires a clear understanding of the intentions of other agents by the time of branching. However, in these works, the probability associated with the uncertain intents of the dynamic agents is assumed to be fixed, and the planning is executed in an \textit{open-loop} fashion. Thus, an exact estimation of the branching time $t_b$ is required, otherwise a collision may occur \cite{Bajcsy}. Learning-based approaches have been recently introduced in the contingency planning context \cite{Rhinehart cont, Packer}, however, they are not interpretable and hard to tune. \cite{Marc} proposes a contingency planning approach that uses a dynamic branching point determined by a predefined heuristic. This heuristic chooses the branching time as the maximum time such that any two future scenarios starting at the current time only diverge by a maximum distance. Despite being effective, this heuristic entails at least double the computational time since the divergence measure is invoked on the ego-vehicle trajectories. 

In this paper, we adopt the idea of splitting the planning problem into short-term and long-term planning.

\subsubsection{\textcolor{black}{Risk-aware motion planning}} \textcolor{black}{Another aspect to consider when planning under uncertainty is to assess the risk associated with the executed plan \cite{Barbosa, Huang, Damerow, Wang, Philipp}. In \cite{Barbosa}, Gaussian process regression is utilized to establish a probabilistic model of the environment. This model is subsequently employed to formulate a risk-aware cost function using the Conditional Value at Risk (CVaR) measure, which is then incorporated into an optimal motion planning algorithm to generate trajectories that avoid high-risk areas. \cite{Damerow} constructs a probabilistic risk map by assessing the anticipated harm considering predicted spatio-temporal trajectories for both the ego-vehicle and other participants in the traffic scene. These maps serve as indicators of the risk associated with a planned trajectory, calculated through a rapidly-exploring random tree algorithm. Despite being effective, a drawback of this approach is that the driven trajectories of the other traffic participants need to be pre-defined and known by the ego-vehicle a priori. Along the same line as our proposed approach, \cite{Philipp, Wang} define risk as a product of two components, the probability of collision with other traffic subjects and the severity of that potential collision. In that sense, an analytic approach is proposed to calculate the probability of spatial overlap of the ego-vehicle with dynamic obstacles at discrete times. }

\subsection{Statement of Contributions}
In this work, we aim to address the state-of-the-art aforementioned limitations. In particular, the contributions of this paper can be summarized in the following points:
\begin{enumerate}[(i)]
\item In contrast to recent contingency planning schemes that assume an open-loop information structure, we propose a Bayesian update scheme that incorporates the observations of the human states into the motion planner cost function, influencing the short-term plan based on the belief the ego-vehicle maintains over human intentions.
\item  We incorporate a probabilistic risk metric into the contingency planner to balance safety and efficiency.
\item We analyze the effect of branching time and the belief over the obstacles'
intents on the short-term plan, and how they relate to the maximum risk the ego-agent endures.
\item We show how the proposed approach can be extended to multi-agent scenarios by leveraging the permutations over all possible intentions the traffic agents can have.
\end{enumerate}

\section{Preliminaries}
In this section, we first introduce the principal tools that constitute the proposed contingency planning framework.
\subsection{Notation}

Throughout this paper, vectors, and matrices are expressed in bold, $\boldsymbol{x}$, and capital bold, $\boldsymbol{A}$, letters respectively. $||\boldsymbol{x}||$ is the Euclidean norm of $\boldsymbol{x}$, and the subscript $(\cdot)_k$ indicates the value at stage $k$,  $f(\cdot)$ is the probability density function. The planning problem is formulated in a receding horizon fashion where only the first control input is executed, and then the whole process is reiterated with the new initial conditions and observations.
\subsection{Ego-Motion Sampler}
\label{sampler}
\begin{figure}[!t]
\centering
\includegraphics[width=2.35in]{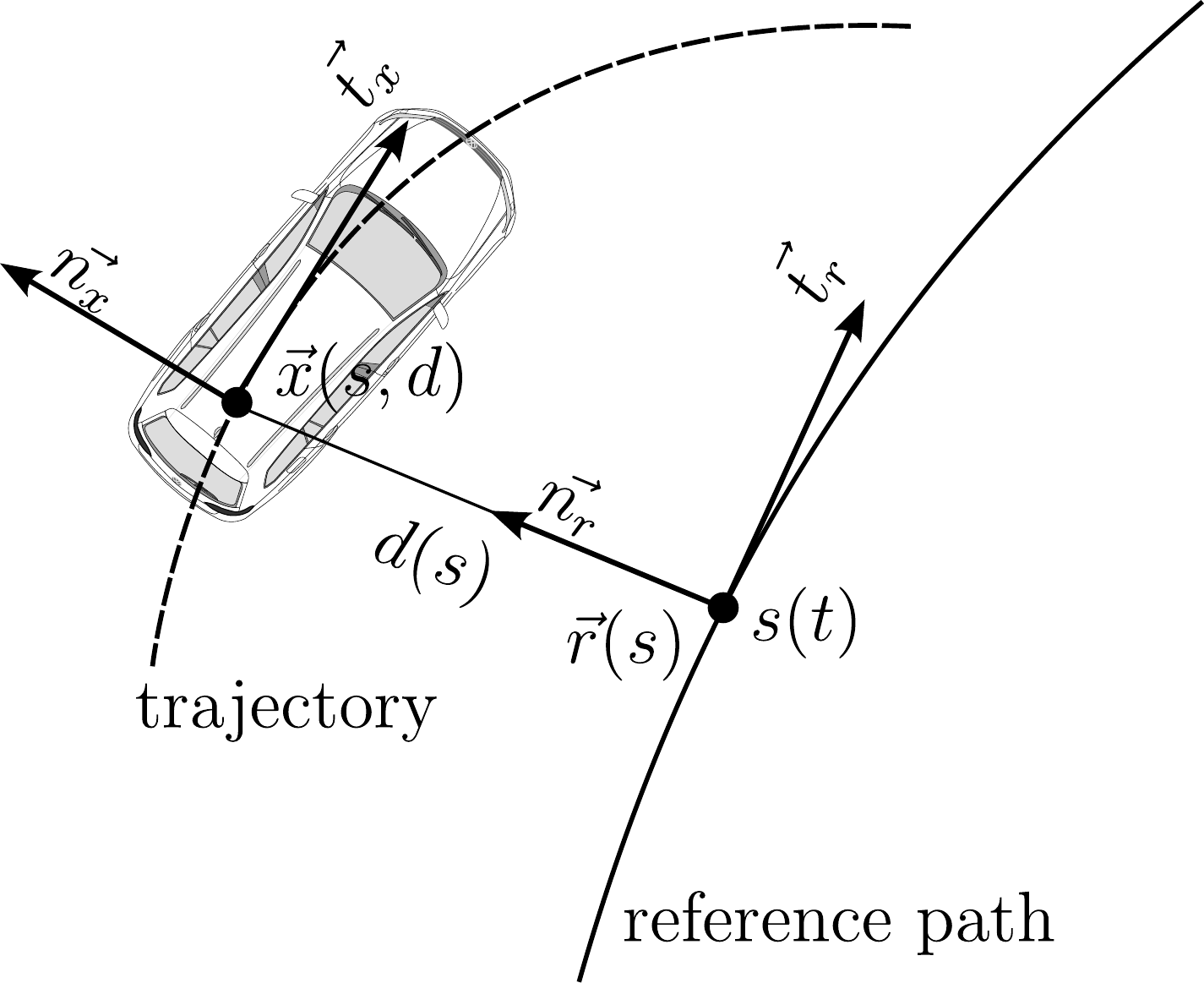}
\caption{Illustration of the Frenet coordinate system.}
\label{Frenet}
\end{figure}

The general form of the planning problem can be formulated as follows:
\begin{subequations}
\begin{alignat}{2}
    \tau^* = & \arg \min_{\tau \in \mathcal{T}_n}J(\tau; \Gamma), \\
    \textrm{s.t.} \quad & g_j (\tau) \leq b_j, \quad j=1, ...,n
\end{alignat}
\end{subequations}
where the optimal trajectory $\tau^*$ is defined as the one giving the minimum total cost $J(\tau)$ from a set of sampled trajectories $\mathcal{T}_n$ given the ego-state. $\tau \in \mathcal{T}_n$ is a continuous path through the state space and is characterized by a sequence of points $\tau = \{\boldsymbol{x}_0, \boldsymbol{x}_1, ..., \boldsymbol{x}_N\}$ defined over a horizon of length $N$ with regular intervals $\Delta t$, where $\boldsymbol{x}_k =(x_k,y_k)$. The optimal trajectory must adhere to a set of time-dependent constraints $g_j(\tau)$ imposed by the surrounding dynamic obstacles, vehicle kinematics, and other user-defined constraints.\\
Instead of formulating the problem directly in the Cartesian coordinate system, we switch to the \textit{Frenet Frame} to exploit the lane-geometric information. In that sense, the trajectory is parameterized by the total arc-length $s(t)$ traveled along the reference path $\Gamma$ parameterized by time $t$, and the orthogonal lateral deviation $d(s)$ parameterized by arc-length $s$ as shown in Fig. \ref{Frenet}. In this paper, we adopt a sampling-based motion planning approach that is widely used in the intelligent vehicles community \cite{Werling, Mutlu, Vosswinkel, Jin, Cheng, FISS}.

To generate a set of possible candidate trajectories, we first need to sample a set of terminal states for both longitudinal and lateral trajectories. To ensure the diversability of the candidate trajectories, it is crucial to emphasize that the sampled terminal states should cover various maneuvers which include maintaining the current velocity, accelerating to a certain speed, yielding velocity profiles for the longitudinal trajectories, and lane-keeping, lane-change, and nudging for lateral trajectories. Given the current ego-state, with respect to the reference path, and the sampled terminal states, piecewise quartic and quintic polynomials can be used to generate the longitudinal and lateral trajectories respectively \cite{Werling}. The Frenet state defined as $\left[s, \dot{s}, \ddot{s}, d, d^\prime, d^{\prime \prime}\right]$ can then be converted to the global coordinates $\left[x, y, \theta, \kappa, v, a\right]$, where $\dot{(\cdot)}:=\frac{\text{d}(\cdot)}{\text{d}t}$, and $(\cdot)^\prime:=\frac{\text{d}(\cdot)}{\text{d}s}$ indicate the parameters derivatives with respect to time and arc-length respectively.

After generating a set of trajectories, some of them are then pruned based on some imposed constraints $g(\tau)$. These constraints are affected by the kinodynamic feasibility of the ego-vehicle, in addition to collision avoidance constraints concerning the surrounding obstacles. Since the motion of the dynamic obstacles is not known a priori, this necessitates the need for a probabilistic prediction model that models their behavior forward in time which is described in Subsection \ref{MoG}. After pruning the invalid trajectories, a user-defined cost function  $J(\tau)$ is assigned to each valid trajectory, and the trajectory $\tau$ with the minimum cost is selected. The details of $g(\tau)$ and $J(\tau)$ formulations are given in Section \ref{problem formulation}. 
 % After a single time step is completed, a new trajectory is planned for the next time step.

\subsection{Probabilistic Prediction Model}
\label{MoG}
The planner is required to stochastically forecast other agents' behavior \textcolor{black}{to make informed decisions. This necessitates the need for a prediction model that can effectively capture the diverse and nuanced intentions exhibited by different drivers. For instance, a neighboring vehicle may opt to either remain in its lane or merge in front of us. To address the inherent uncertainty and complexity associated with the multi-modal behavior of road users, the application of Mixture-of-Gaussians (MoG) distributions is widely used in literature, see e.g., \cite{Chai, Hong}. It is important to highlight that other multi-modal prediction methods could also be used in
conjunction with our proposed approach.} In our settings, there exists a set of dynamic obstacles $o \in \mathcal{I}_o := \{1,...,n_o\}$, at position $\boldsymbol{\delta}^o \in \mathbb{R}^2$. The probability measure associated with the uncertainty of the perception of the dynamic obstacles is denoted by $\mathbb{P}$ and defined over the probability space $\Delta$. \textcolor{black}{An MoG} model serves as a comprehensive method for articulating uncertainty characterized by multiple modes, achieved through the integration of multiple continuous probability distributions,
\begin{equation}
    f_k^o(x,y) = \sum_{i=1}^{n}\phi_i f_{k,i}^o(x,y),
    \label{pdf}
\end{equation}
where $n$ is the number of modes of the MoG, $\phi_i$ represents the weight of each mode such that $\sum_{i=1}^n \phi_i= 1$, and $f_{k,i}^o(\cdot)$ is the probability density function of each mode with mean $\boldsymbol{\mu}_i \in \mathbb{R}^n$ and covariance $\boldsymbol{\Sigma}_i \in \mathbb{R}^{n \times n}$,
\begin{equation}
    f_{k,i}^o(\cdot) = \mathcal{N}(\boldsymbol{\mu}_i(t), \boldsymbol{\Sigma}_i(t)),
\end{equation}
The output of the prediction model is subsequently a sequence of predicted state distributions along the prediction horizon for a given mode. This will be later leveraged to calculate the risk associated with the planned trajectories as detailed in the following subsection.
\subsection{Collision Chance Constraints}
The ego-vehicle $v$ and the obstacle $o$ are mutually collision-free if 
$
||\boldsymbol{x}^v_k - \boldsymbol{\delta}_k^o|| \leq r
$, where $\boldsymbol{x}^v, \boldsymbol{\delta}^o \in \mathbb{R}^2$ denote the positions of the ego-vehicle and obstacle respectively, and $r$ is the safety distance. However, since the positions of the obstacles are defined as random variables, the collision avoidance constraints can only be satisfied in a probabilistic manner, and thus defined as chance constraints at each timestep $k$,
\begin{equation}
  \mathbb{P}(||\boldsymbol{x}^v_k - \boldsymbol{\delta}_k^o|| \leq r) \leq 1 - \delta 
  \label{distribution}
\end{equation}
where $\mathbb{P}$ indicates the probability measure, and $\delta \in (0,1]$ is the collision probability threshold. 

\section{Problem Formulation}
In traditional motion planning frameworks, a single trajectory is sought to minimize the expected cost over all plausible predicted futures along the entire planning horizon \cite{Xu}, which may result in sub-optimal and overly-conservative trajectories. The contingency planning paradigm, on the other hand,  generates a distinct set of trajectories conditioned on the different outcomes from the prediction model.

\subsection{\textcolor{black}{Contingency Planning}}
Due to the stochastic nature of the surrounding agents' intentions, contingency planning outputs multiple policies $\Pi$ where each policy $\pi \in \Pi$ is specified for a single agent's intent $\tilde{\lambda}$. Given the ego-vehicle's incapacity to concurrently traverse multiple contingency plans, the initial segment of each plan  $\tau_{0:t_b}$ is restricted to remain consistent. The contingency planning problem can be formulated as,
\begin{subequations}
\begin{alignat}{2}
    \arg & \min_{\tau}J_\text{shared}(\tau_{0:t_b}) +  \sum_{\lambda \in \Lambda} p(\lambda) J_\text{conting} (\tau_{t_b:T}, \lambda) \\
    & \textrm{s.t.} \quad  g_j (\tau) \leq b_j, \quad j=1, ...,n
\end{alignat}
\end{subequations}
where $J_{\text{shared}}$ is the cost of the shared part of the plan that takes into consideration all possible modes of the prediction model, $t_b$ is the branching time representing the time at which the shared plan bridges into different contingent plans, $J_\text{conting}$ is the cost associated with each contingent plan, and $p(\lambda)$ indicates the probability of each possible intention given by the prediction model introduced in Section \ref{MoG}.\\
However, with this problem formulation, the following challenges arise:
\begin{enumerate}[(i)]
\item  It is assumed that by $t_b$, the uncertainty about the other agents' intentions is resolved, and the ego-vehicle branches to the predicted true hypothesis. Thus, $t_b$ has to be calculated accurately, otherwise, the ego-vehicle may choose the wrong branch which can result in a collision.
\item The hypothesis probabilities estimated by the prediction model, $p(\lambda)$, are usually of a bad quality and entirely relying on them could result in collisions \cite{Nakamura}.
\end{enumerate}

In \cite{Bansal, Bajcsy2, Bajcsy}, an offline reachability analysis approach is proposed to alleviate the first issue by estimating the branching time $t_b$. However, this approach requires a discretization of the state space and takes into consideration the worst behavior of the other agents which leads to the worst-case estimate of $t_b$, that is the latest time at which the ego-vehicle becomes certain about the other obstacle's intention. Moreover, due to the curse of dimensionality, this approach can barely extend to multi-agents.

Therefore, in our proposed approach, introduced in the upcoming section, we tackle the first issue by introducing a belief updater that updates the ego-vehicle's prior belief about obstacles' intentions based on the online measurement it perceives. Moreover, to address the inherent trade-off between safety and efficiency, we introduce risk-aware contingency planning by augmenting contingency planning with a probabilistic risk measurement. 
\section{Proposed Approach}
\label{problem formulation}

\subsection{Planning under Uncertain Intentions}
To alleviate the conservatism of the planned trajectory, we propose to have a probabilistic inference over the possible intents that an obstacle can have by introducing a Bayesian belief updater instead of solely relying on the prediction model to make an informed estimate of the unknown intent.
\subsubsection{Belief Updater}
 At each time step, we assume that the ego-vehicle can observe the true state $\hat{\boldsymbol{\delta}}_t^o$ of the dynamic obstacle, but not its internal state. This enables the ego-vehicle to retrospectively assess the likelihood of the dynamic obstacle's observed states under the prediction model. Thus, the ego-vehicle always maintains a belief, $b(\lambda)$, over the set of possible intents $\Lambda$ of other agents, i.e., $\lambda \in \Lambda$. A \textit{Bayesian filter} is used to update the ego vehicle's belief over the obstacle's intents based on the new observations the ego-vehicle perceives. The update rule for the obstacle's intent is given by,
\begin{equation}
    b(\lambda)_+^t=\frac{f(\hat{\boldsymbol{\delta}}_t^o; \boldsymbol{\mu}_t, \boldsymbol{\Sigma}_t) b(\lambda)_-^t}{\sum_{\tilde{\lambda}}f(\hat{\boldsymbol{\delta}}_t^o; \boldsymbol{\mu}_t, \boldsymbol{\Sigma}_t)b(\tilde{\lambda})_-^t}
    \label{belief}
\end{equation}
where $b(\tilde{\lambda})_-^t$ represents the prior belief of the ego-vehicle on the intent $\tilde{\lambda}$ at time $t$. $b(\tilde{\lambda})_+^t$ indicates the same probability a posteriori. $f(\hat{\boldsymbol{\delta}}_t^o; \boldsymbol{\mu}_t, \boldsymbol{\Sigma}_t)$ denotes the probability density function of a single mode of the MoG, given by the prediction model defined in Section \ref{MoG}, with mean $\boldsymbol{\mu}_t$ and covariance $\Sigma_t$, and evaluated at the observed state $\hat{\boldsymbol{\delta}}_t^o$.
\begin{rmk}
    \textit{Based on the provided prediction model and what the ego-agent can observe, the probability density function $f(\cdot)$ can either be defined on the obstacle's state or the control input}.
\end{rmk}

\subsubsection{Multi-Agent Scenario} In a multi-agent setting, it is not sufficient to only consider the belief that the ego-agent maintains over a single agent's intentions, but rather to consider how the traffic scene would evolve as a whole into the future. To address this issue, it is required to consider all the permutations $\Theta = \{\theta_1,...,\theta_{n_s}\}$ where $n_s$ is the total number of realizations the traffic scene can evolve to, and $\Theta$ is determined by the Cartesian product of all obstacles policies, $\Theta = \Lambda_1 \times ... \times \Lambda_{n_s}$. The total number of realizations, $n_s$, is defined by the cardinality of $\Theta$, $n_s = |\Theta|$. The probability of each realization can, subsequently, be calculated by,
\begin{equation}
    p(\theta_j)_+ = \frac{p(\theta_j)_- \prod_{i=1}^{n_o} b_i(\lambda)_+^t}{\sum_{j=1}^{n_s} p(\theta_j)_- \prod_{i=1}^{n_o}b_i(\lambda)_+^t}
    \label{permutation}
\end{equation}
where $b_i(\lambda)_+^t$ is determined by \eqref{belief}, and $\lambda$ is the corresponding intention for obstacle $i$ that belongs to $\theta_j$. $p(\theta_j)_+$ is used as a weight for the contingent plans in the cost function defined in \eqref{10a}. This mimics a scene-centric prediction model by outputting modes of joint trajectories with respect to all agents in the scene.
\subsubsection{Probabilistic Risk Assessment}
In our proposed approach, to achieve probabilistic collision avoidance, we rely on \textcolor{black}{an existing \textit{risk metric}, motivated by our previous work \cite{Mustafa},} that maps the distribution of a random variable, defined in \eqref{distribution}, to a real number. 
\begin{definition} [Risk Metric]  \textit{Let} $ \mathcal{Z} $ \textit{denote the set of random variables representing the uncertainty of the obstacles' motion in the} $x$ \textit{and} $y$ \textit{directions. The risk metric maps the distribution of the random variables to a real number indicating \textcolor{black}{the induced risk,}} \textcolor{black}{$\mathcal{R}$}$: \mathcal{Z} \mapsto \mathbb{R}$.
\end{definition}

\textcolor{black}{There exist various definitions of risk in the literature. Among them, safety standards, \cite{ISO1, ISO2} define risk as a combination of the probability of occurrence of harm and the severity of that harm. Inspired by this definition, different approaches in robotics and autonomous driving community \cite{Ethical, Mustafa, Trolley, Wang} define risk as the product of two quantities, namely the probability of collision that the planned trajectory has with any of the surrounding obstacles and the level of severity linked to that potential collision at every time-step, $k$, along the planning horizon,
\begin{equation}
   \mathcal{R}_k^o (\boldsymbol{x}_k^d) = \mathcal{C}_k^{o}(\boldsymbol{x}_k^d)\mathcal{S}_k^{o}(\boldsymbol{x}_k^d), \quad \forall k,o,d
\end{equation}
Here, we approximate the ego-vehicle by two discs, where each disc is referred to by $d$, and calculate the probability of collision for each disc. Note that we model the ego-vehicle as two discs such that we can consider all its shape in calculating the induced risk. Given the probability density function of the prediction model as indicated in \eqref{pdf}, the probability of collision, for each ego-vehicle's disc ate each planning stage $k$ per obstacle $o$}, can be calculated by estimating the spatio-temporal overlap of the predicted modes with the ego-vehicle's plan $\tau$, 
\begin{equation}
    \mathcal{C}_k^{o}(\boldsymbol{x}_k^d) =
    \iint_{x^d_k, y^d_k \in D}  f_k^{o}(x,y) dx dy, \quad \forall{k}, o, d,
\end{equation}
which is an integral of the Mixture of Gaussians (MoG) probability density function over a specified domain where the integration domain $D$ is defined as a circle whose center is located at the predicted vehicle pose $\boldsymbol{x}_k^d$ at stage $k$ along the prediction horizon, and its radius $r$ is the sum of the vehicle and obstacle radii since the PDF does not account for the obstacle-vehicle's shape. 

Another aspect of the risk assessment is the determination of the severity of a potential collision for each planned trajectory. \textcolor{black}{The expected collision severity is of high importance in case a collision is inevitable and the best behavior has to be selected to reduce upcoming damage. The collision severity definition is motivated by the work proposed in \cite{Ethical, Wang} which can be determined, $\forall k, o, d$, by} 
\begin{equation}
 S_k^{o}(\boldsymbol{x}_k^d) =
     \frac{\text{m}^v}{\text{m}^v+\text{m}^o}((v_k^v)^2+(v_k^o)^2-2v_k^vv_k^o\cos\alpha)^{\frac{1}{2}}
    \label{severity}
\end{equation}
where $\text{m}^v$ and $\text{m}^o$ represent the masses of the ego-vehicle and obstacle vehicle respectively whereas $v^v$ and $v^o$ are their corresponding velocities, and $\alpha$ is the collision angle. \textcolor{black}{Here, it is important to emphasize that establishing an appropriate severity measure is a challenging problem and constitutes a dedicated area of research that is beyond the scope of this paper. Notably, the severity metric outlined here does not account for ethical considerations associated with the resultant damage, including the vulnerability of road users. This can, nevertheless, be embedded in \eqref{severity}, in case semantic information of the road users is provided by a perception model. In this context, leveraging semantic categorization allows for the scaling of the severity metric within a range of 0 to 1. A severity value of 0 corresponds to a collision causing no harm, while a severity value of 1 signifies the highest level of damage, particularly involving vulnerable road users. This scaled metric can then be incorporated into the sampling-based planner to discard trajectories that lead to collisions with vulnerable road users.}

Furthermore, we adopt a discounted chance constraint formulation, as proposed by \cite{Yan}, where violation probabilities close to the early stages of the plan have higher penalization compared to the ones in the far future. The discounted \textcolor{black}{risk} at every stage $k$ concerning an obstacle $o$ is defined as
\begin{equation}
   \mathcal{R}_k^o (\boldsymbol{x}_k^d) = (\gamma)^k \mathcal{C}_k^o(\boldsymbol{x}_k^d)\mathcal{S}_k^{o}(\boldsymbol{x}_k^d), \quad  \forall k, o, d
\label{discounted probability}
\end{equation}
where $\gamma \in (0,1]$ is the discounting factor.

 After calculating the risk for each ego-vehicle's disc at every timestep $k$ along the planning horizon per each obstacle $o$, the predicted risk $\eta$ at the current time step is defined by maximizing risk for all obstacles, $o\in \mathcal{I}_o$, at all planning horizon stages $k$ and all discs $d$,
\begin{equation}
   \eta = \max_{o \in \mathcal{I}_o, k, d} \mathcal{R}_k^o (\boldsymbol{x}_k^d),  \label{max_risk}
\end{equation}

The max operator in \eqref{max_risk} ensures that the worst-case risk over the planned trajectory is below the threshold $\delta$. 

\begin{rmk}
    \textcolor{black}{\textit{According to \cite{coherent risk}, the quantified risk metric given in \eqref{discounted probability} is non-coherent since it does not fulfill all coherence axioms. Nevertheless, it is efficient in capturing the underlying uncertainty associated with the motion of the dynamic obstacles and scales monotonically as the level of risk increases. Here
it should be emphasized that, since our planner is sampling-based, it is agnostic to the risk metric
employed and a comparative study on different risk metrics will be part of future research.}}
\end{rmk}

Thus, the optimization problem can be formulated as follows,
\begin{subequations}
\begin{alignat}{2}
    \min_{\tau \in \mathcal{T}} \quad & J_{\text{shared}}(\tau_{0:t_b})+\sum_{j=1}^{n_s} p(\theta_j)_+J_{\text{conting}}(\tau_{t_b:T}(\theta_j)) \label{10a} \\
     \textrm{s.t.} \quad  & g_j (\tau) \leq b_j, j=1, ...,n\\
     & p(\theta_j)_+ = \frac{p(\theta_j)_- \prod_{i=1}^{n_o} b_i(\lambda)_+^t}{\sum_{j=1}^{n_s} p(\theta_j)_- \prod_{i=1}^{n_o}b_i(\lambda)_+^t}, \\
     & \max_{o \in \mathcal{I}_o, k, d} \mathcal{R}_k^o (\boldsymbol{x}_k^d) \leq \delta
\end{alignat}
\end{subequations}
It is worth mentioning that the number of planned contingency trajectories is determined by the number of modes of the prediction model, $|\Lambda|$.

 \begin{figure*}
\centering
\begin{subfigure}{.33\textwidth}
  \centering
  %\hspace{-1.0cm}
  \includegraphics[width=0.95\linewidth]{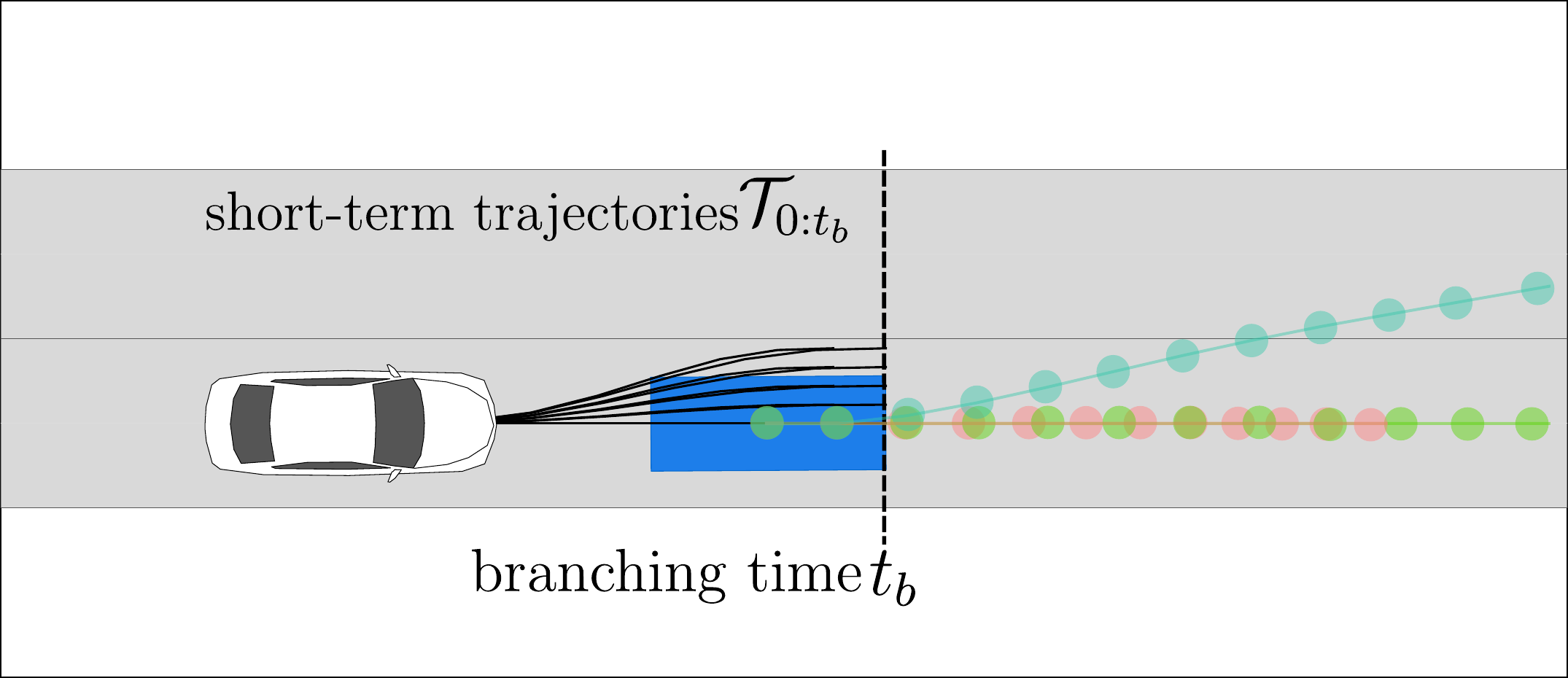}
  \caption{\textcolor{black}{Sampling short-term trajectories till $t_b$}.}
  \label{sub1}
\end{subfigure}%
\begin{subfigure}{.33\textwidth}
  \centering
  \includegraphics[width=0.95\linewidth]{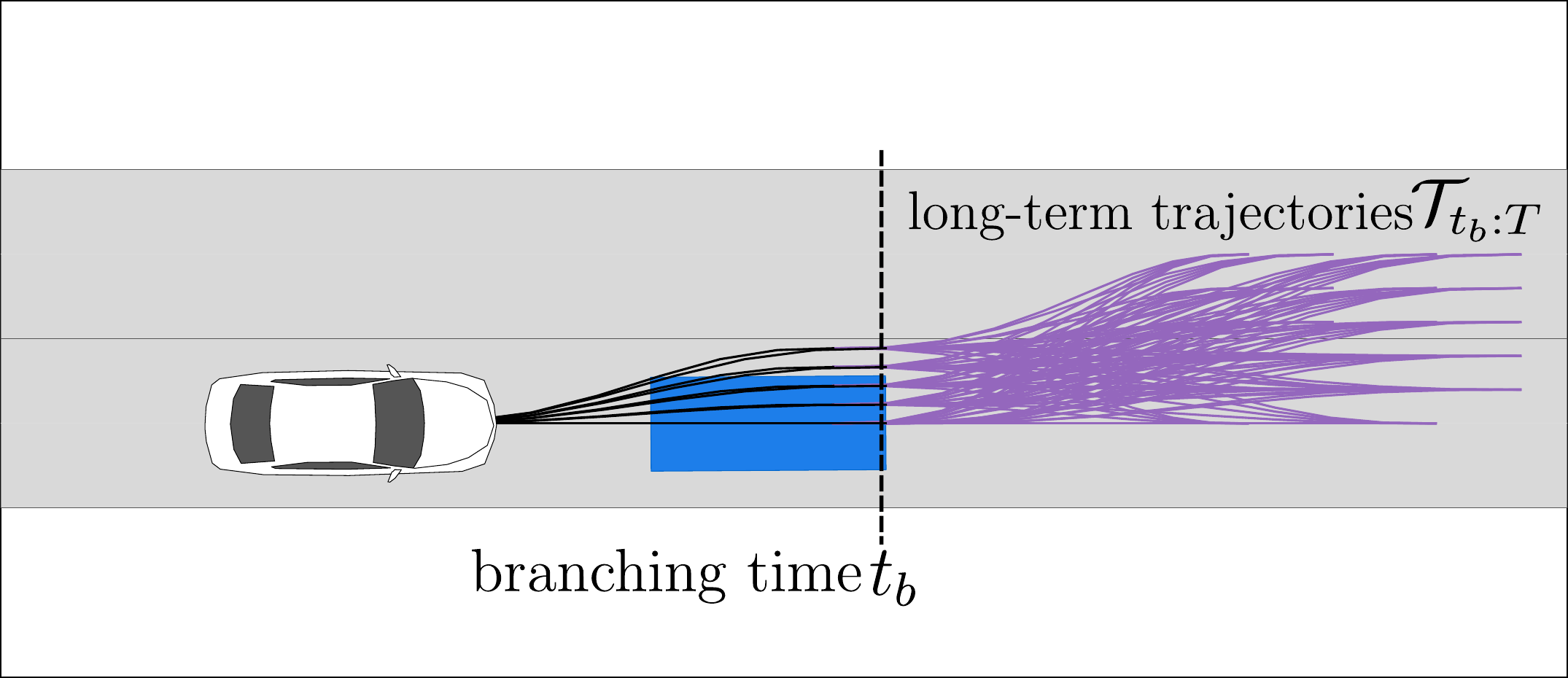}
  \caption{\textcolor{black}{Sampling long-term trajectories till $T$.}}
  \label{sub2}
\end{subfigure}
\begin{subfigure}{.33\textwidth}
  \centering
  \includegraphics[width=0.95\linewidth]{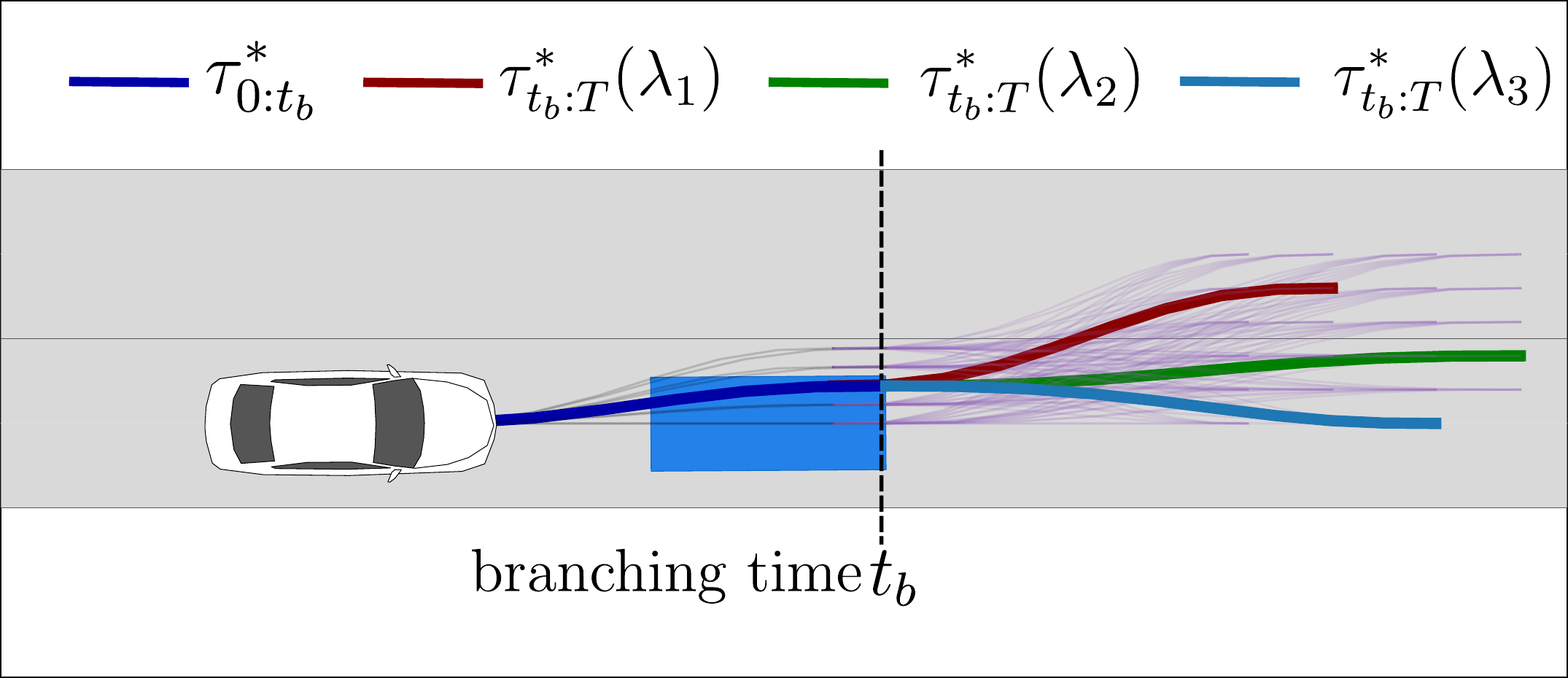}
  \caption{\textcolor{black}{Select the optimal total plan. }}
  \label{sub3}
  %\hspace{-3.0cm}
\end{subfigure}
\caption{Snapshots from the CommonRoad simulation framework for an overtaking scenario where the ego-vehicle is depicted by the car icon and the obstacle-vehicle is illustrated as a blue rectangle. The obstacle-vehicle has three policies that it can execute, $ \Lambda = \{\texttt{maintain speed}, \texttt{slow down}, \texttt{lane change}\}$, where the obstacle-vehicle's predicted trajectories are illustrated in distinct colors in \ref{sub1}. We first sample a set of short-term plans, $\mathcal{T}_{0:t_b}$ visualized in black, till the branching time $t_b$ shown in \ref{sub1}. We then sample a set of long-term plans, $\mathcal{T}_{t_b:T}$ depicted in purple, conditioned on the terminal states of the short-term plans, shown in \ref{sub2}. For the long-term plans, a cost is assigned to each plan per obstacle-vehicle policy $\lambda_i$. In Fig. \ref{sub3}, the total cost of the entire plan is computed by \eqref{10a}, and the total optimal plan is the one given by $ \tau^* = \tau^*_{0:t_b} \cup \{\tau^*_{t_b:T}(\lambda_1), \tau^*_{t_b:T}(\lambda_2), \tau^*_{t_b:T}(\lambda_3)\}$. In this example, the belief over the first policy is dominant, $b(\lambda_1) = 0.63$, compared to the rest causing the optimal short-term plan $\tau^*_{0:t_b}$ to be more biased towards $\tau^*_{t_{b:T}}(\lambda_1)$.}
\label{fig:branching_topology}
\end{figure*}

\subsubsection{Cost Function}
The cost function $J(\tau)$ consists of sub-costs that focus on different aspects of the plan's performance such as safety, passenger comfort, progress, and tracking. It is, thus, defined as,
\begin{equation}
    J(\tau) = \boldsymbol{w}^T c(\tau; \Gamma)
    \label{cost}
\end{equation}
where the weight vector $\boldsymbol{w} \in \mathbb{R}_+ $ captures the weights associated with each cost term. These costs include, for instance, how much the final point of the planned trajectory deviates from the reference path $c_d=|d-d_\text{ref}|$, $c_v=|\dot{s}-\dot{s}_\text{ref}|$ is the deviation from the reference velocity, $c_p = \sum_{k=0}^{N-1}||\boldsymbol{x}^v_{k+1} - \boldsymbol{x}^v_k|| $ is the cost representing progress, i.e., the total traveled distance along the reference path $\Gamma$, and $c_j = \sum_{k=0}^N\frac{\dddot{s_k}^2+d^{\prime \prime \prime2}_k}{N}$ penalizes the mean square sum of longitudinal and lateral jerks used as a comfort indicator.\subsubsection{Kinematic Constraints}
\textcolor{black}{To guarantee a smooth and comfortable transition between the short-term shared plan and the contingency plans, when concatenated,} we need to impose some constraints on the curvature at the branching point. From \cite{Cheng},
\begin{equation}
    d^{\prime \prime}= -\left[\kappa_rd\right]^\prime \tan \theta + \frac{1-\kappa_rd}{\cos^2\theta} \left[\kappa_p \frac{1-\kappa_rd}{\cos \theta} - \kappa_r\right],
\end{equation}
where $\kappa_r$ and $\kappa_r^\prime$ denote the curvature of the reference path $\Gamma$ and its derivative respectively. By rearranging terms, an explicit formulation of the trajectory's curvature can be obtained,
\begin{equation}
    \kappa_p = \left[d^{\prime \prime}-(\kappa_rd)^\prime \tan \theta\right]\frac{\cos^3\theta}{(1-\kappa_rd)^2}+\frac{\kappa_r \cos \theta}{1-\kappa_rd}
\end{equation}
After calculating the curvature and every point along the trajectory, a box constraint is defined as
\begin{equation}
    |\kappa_p| \leq \kappa_{\text{max}}
\end{equation}
where $\kappa_{\text{max}}$ at every timestep is parameterized by the planned velocity of the ego-vehicle and the maximum allowable lateral acceleration. Finally, to ensure that the planned trajectory is kinematically feasible, the following box constraints must be satisfied at every timestep,
\begin{subequations}
\begin{alignat}{2}
   v_{\text{min}} \leq v(t) \leq v_{\text{max}} \\
      a_{\text{min}} \leq a(t) \leq a_{\text{max}} \\
            j_{\text{min}} \leq j(t) \leq j_{\text{max}}
\end{alignat}
\end{subequations}
where $v_{\text{min}} / v_{\text{max}}$ represent the minimum and maximum velocity, $a_{\text{min}} / a_{\text{max}}$ represent the minimum and maximum acceleration, and $j_{\text{min}} / j_{\text{max}}$ represent the minimum and maximum jerk.

\begin{figure}[!t]
\centering
\includegraphics[width=2.8in]{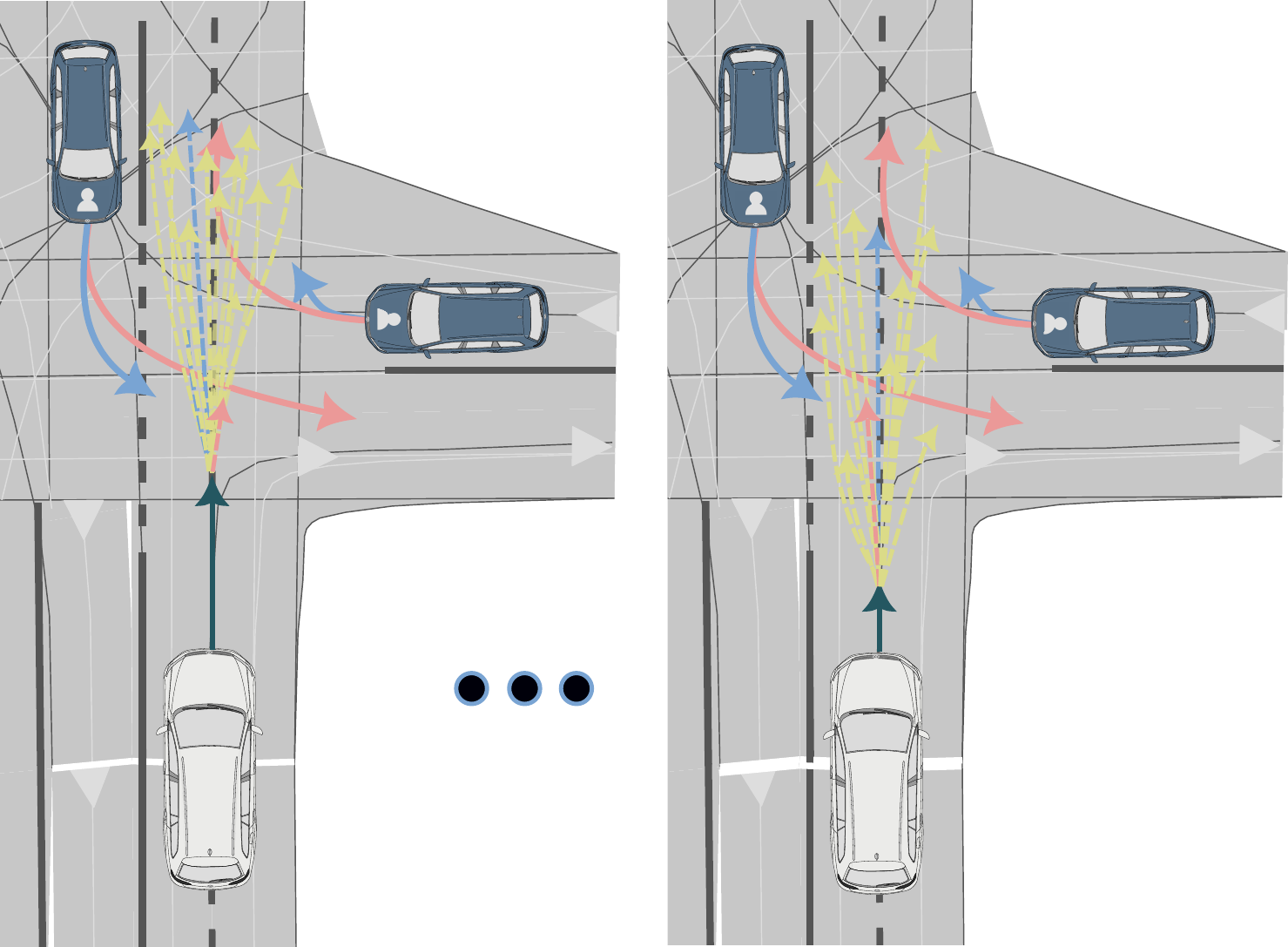}
\caption{Contingency planning paradigm. A large set of short-term plans are sampled together with multiple long-term plans. The total plan with the minimum cost for each possible future realization is selected (indicated in blue and red).}
\label{fig_3}
\end{figure}
\subsection{Planner Design}
To better explain the proposed approach, we recall the illustrative example in which the ego-vehicle is not fully aware of the future trajectories of another agent or even its intended goal, as illustrated in Fig. \ref{fig_3}. In this example, we take into consideration only two likely futures. The one depicted in blue is where the other agent yields to the ego-agent, and the one in red is where the ego-agent must brake before entering the intersection since the oncoming vehicle is taking an aggressive left turn.\\
\begin{algorithm}
\caption{Risk-aware contingency planning}\label{alg:cap}
\begin{algorithmic}[1]
\Require Prediction model per obstacle $f_{k,i}^o(\cdot)$, horizon $T$, branching time $t_b$, risk tolerance level $\delta$, reference path $\Gamma$, prior belief per intention $\lambda$, $b(\lambda)_-=\frac{1}{|\Lambda|}$, two priority queues for sorting candidate trajectories $Q, Q_\text{final}$.
\Ensure{ An optimal shared plan $\tau_{0:t_b}$, concatenated with $|\Lambda|$ contingent plans $\mathcal{T}_{t_b:T}$}.
\ForAll{$t=1,2,...$}
\State $\mathcal{T}_{0:t_b} \gets \text{SampleSharedTrajectories$(\boldsymbol{x}_{\text{init}})$}.  $

\ForEach {$\tau_{0:t_b} \in \mathcal{T}_{0:t_b}$}
\State   $\eta \gets \text{EvaluateRisk}(\boldsymbol{x}_k^d, \boldsymbol{\delta}_k^v), \forall k, v, d $ \eqref{max_risk}.
   \If {$\eta \geq \delta$}
   \State \textbf{continue}
      \EndIf
\State   \textcolor{black}{$J_{\text{shared}}(\tau_{0:t_b}) \gets \text{ComputeCost}(\boldsymbol{w}, \tau_{0:t_b}, \Gamma)$ \eqref{cost}.}
   \ForEach {$\lambda \in \Lambda$}
     \State $\mathcal{T}_{t_b:T} \gets \text{SampleContingentPlans($\tau_{0:t_b}(\text{end})$).}  $
     \State $Q=\{\}.$
   \ForEach {$\tau_{t_b:T} \in \mathcal{T}_{t_b:T}$}
   \State   $\eta \gets \text{EvaluateRisk}(\boldsymbol{x}_k^d, \boldsymbol{\delta}_k^v, \lambda), \forall k, v, d$.
      \If {$\eta \geq \delta$}
   \State \textbf{continue}
      \EndIf
         \State   $J_{\text{cont}}(\tau_{t_b:T}) \gets \text{ComputeCost}(\boldsymbol{w}, \tau_{t_b:T}, \Gamma)$.
         \If {$\tau_{t_b:T}$ passed constraint check}
\State add $\tau_{t_b:T}$ to $Q$.
\EndIf
   \EndFor
   \State  $\tau_{t_b:T}(\lambda)_\text{best} \gets$ pop the first candidate from $Q$.
      \State $b(\lambda)_+ \gets \text{UpdateBelief}(f_{k,i}^o(.), b(\lambda)_-)$ \eqref{belief}.
\EndFor  
   \State $ \tau = \tau_{0:t_b} \cup \{\tau_{t_b:T}(\lambda_1)_\text{best}, \ldots, \tau_{t_b:T}(\lambda_{|\Lambda|})_\text{best}\}$.
   \State $p(\theta_j)_+ \gets$ \text{UpdatePermutations}($b_i(\lambda)_+$) \eqref{permutation}.
      \State Evaluate total cost using \eqref{10a}.
      \State add $\tau$ to $Q_\text{final}$.
\EndFor
\State $\tau_\text{best} \gets$ pop the first candidate from $Q_\text{final}$.
\EndFor
\end{algorithmic}
\end{algorithm}
The proposed approach is summarized in Algorithm \ref{alg:cap}:
\begin{enumerate}[(i)]
\item In lines (2-8), we sample a large set of shared plans $\mathcal{T}_{0:t_b}$ using the ego-motion sampler explained in section \ref{sampler}. This is followed by pruning trajectories that violate the risk upper limit or kinematic constraints. Here, the risk analysis is done with respect to all prediction modes.

\item Line 9 iterates over the number of modes coming from the prediction model.

\item In lines (10-21), for each shared plan, conditioned on its end state, we sample a set of long-horizon trajectories $\mathcal{T}_{t_b:T}$. A risk analysis is performed for each long-term trajectory with respect to a single mode from the prediction model, and trajectories that violate constraints are pruned. A cost is assigned to each trajectory according to \eqref{cost}. 
\item Lines (22-23) sort the long-term trajectories based on their associated costs and pick the trajectory with the minimum cost. The belief over the prediction mode, we iterate over, is updated using \eqref{belief}.
\item In lines (25-27), the shared-trajectory is concatenated with all contingent plans. The total cost of the entire plan is computed by the expected cost of the contingent plans in addition to the cost of the shared plan itself as indicated in \eqref{10a}. 
    \item We find the optimal response, for each shared plan, associated with each scenario from the corresponding long-horizon trajectories in line 30.

\end{enumerate}
This process is then repeated at every time step in a receding horizon manner.
\begin{rmk}
\label{invalid}
    \textit{In case no valid trajectory is obtained from the set of candidate trajectories $\mathcal{T}$, we apply the trajectory with the least risk as long as it is dynamically feasible.}
\end{rmk}

\begin{figure}[!t]
\centering
\includegraphics[width=2.8in]{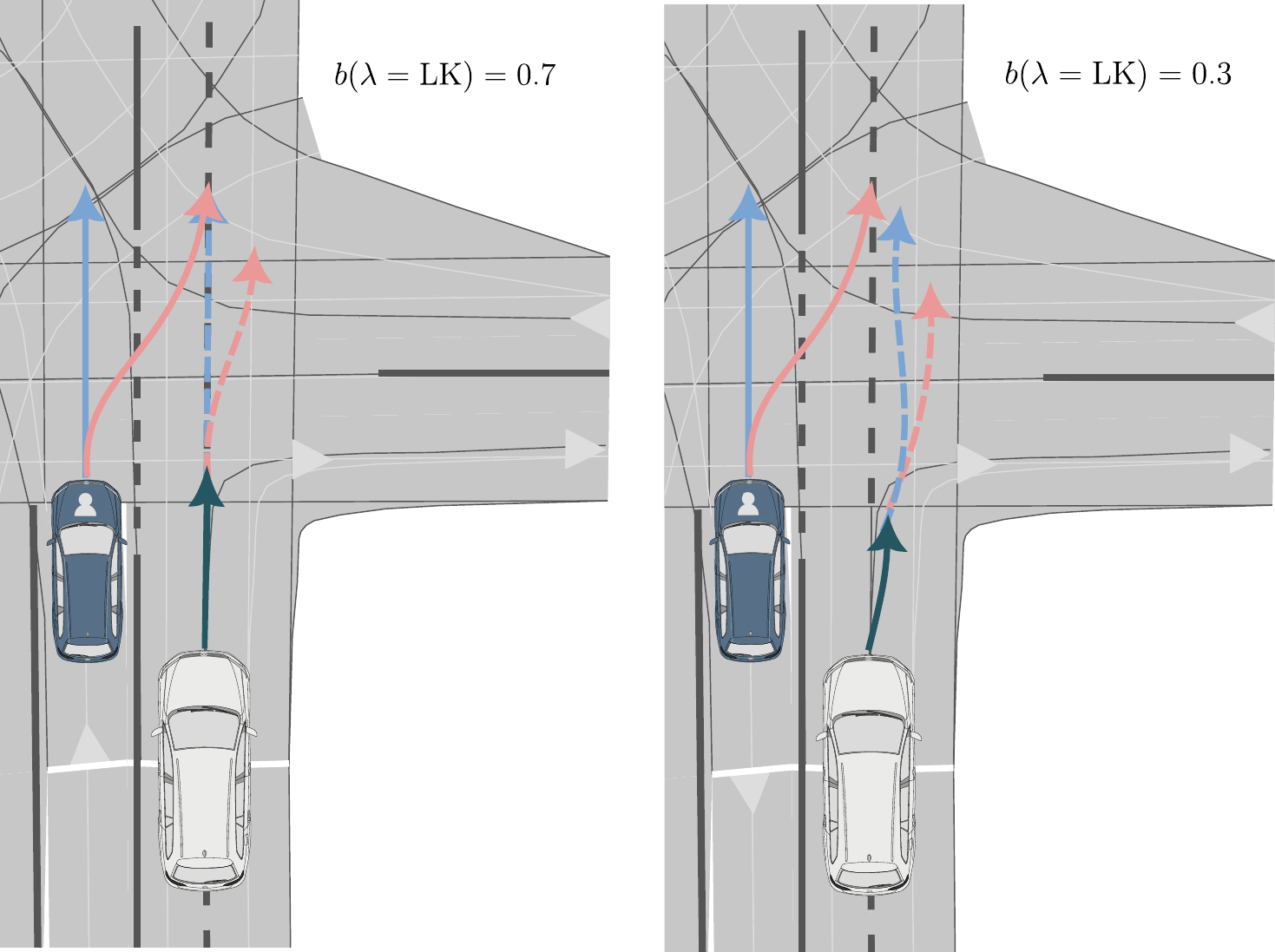}
\caption{The human-driven vehicle can have two different policies, lane-keeping (LK) or lane-change (LC). On the left, the ego-vehicle has a higher belief that the human-driven vehicle executes the LK policy. That's why the short-term trajectory tends to accelerate. On the right, the ego-vehicle has a higher belief that the human-driven vehicle executes the LC policy causing the short-term plan to decelerate and steer to the right.}
\label{jaywalking}
\end{figure}

\textcolor{black}{Here it is crucial to emphasize that an advantage of using a sampling-based planner in the Frenet frame as a basis for our contingency planning framework is that the sampled trajectories cover various maneuvers that the ego-vehicle can have. Thus, in contrast to the approaches proposed in \cite{Chen, Oliveira, Fors}, our approach eliminates the necessity for a pre-established trajectory tree or branching topology to formulate contingent plans. Subsequently, the ego-vehicle is not confined to a specific set of predefined policies. This concept is further explained in Fig. \ref{fig:branching_topology} where we show how the optimal plan is selected from the sampled short-term and long-term trajectories. Additionally, it illustrates how the short-term plan is shaped based on the belief the ego-vehicle maintains over the obstacle-vehicle policies.}
Fig. \ref{jaywalking} shows how the belief the ego-vehicle maintains over the long-term plans affects the behavior of the short-term plan. It is observed that when the ego-vehicle has a higher belief in one of the human's intents, the shared plan tends to be biased towards the corresponding contingent plan. In particular, as illustrated in Fig. \ref{jaywalking}-left, when the ego-agent has a higher belief that the human-driven vehicle aims at executing the lane-keeping policy, the shared-plan biases its motion to accelerate along its lane. On the other hand, when the higher belief is assigned to the lane-change maneuver, the shared-plan tends towards steering a bit to the right while decelerating. This is achieved by balancing the cost of the shared-plan according to the likelihood of the beliefs. 
\begin{rmk}
  \textit{It should be pointed out that, in the human-driven vehicle's lane-keeping policy, we still generate a contingent plan for the other possible intent after the branching point, allowing the ego-vehicle to smoothly steer to the right while decelerating in case the human-driven decides to execute the lane-change policy.}  
\end{rmk}

\section{Results}
\label{results}
\subsection{Experimental Setup}
We evaluate our approach in two safety-critical simulated scenarios inspired by autonomous driving interactions. The illustrated scenarios are included in the CommonRoad benchmark suite \cite{CommonRoad} for reproducibility. The first scenario highlights the reactive behavior that the ego-vehicle's plan induces on an obstacle-vehicle in an overtaking scenario. \textcolor{black}{Four different baselines are introduced to compare our proposed contingency planning with. \\
\textbf{Baseline 1: Multi-policy planning} \cite{Chen}, this baseline uses a branch-MPC whose objective is to minimize the expected cost across all branches within a trajectory tree. \\
\textbf{Baseline 2: Robust baseline} \cite{FISS+} that optimizes a single trajectory along the planning horizon that is robust with respect to all predicted modes regardless of their probabilities.\\
\textbf{Baseline 3: Maximum-likelihood estimate} \cite{Xu}, that only considers the most probable mode given by the prediction model while ignoring the rest.\\
\textbf{Baseline 4:} Similar to \cite{Zhan}, we use the mode probabilities, provided by the prediction model, directly in the contingency planning cost function instead of the beliefs obtained from the belief updater. This baseline is used to analyze the effect of the Bayesian belief updater on the contingency planner's performance.}

In contrast to the branch-MPC approach  \cite{Chen} which encounters scalability challenges when addressing multiple obstacle vehicles, primarily due to its exponential complexity requiring a pruning protocol, our proposed approach exhibits seamless adaptability to multi-vehicle scenarios. This is illustrated in the T-junction, and intersection scenarios, where the ego-vehicle interacts with multiple vehicles whose intents are not known to the ego-vehicle a priori\footnote{A video of the simulated experiments accompanies this paper.}. The computer running the simulations is equipped with an Intel\textsuperscript{\textregistered} Core\textsuperscript{TM} i7 CPU@2.6GHz. 
\subsection{Scenario 1. Overtaking in Highway Driving:} 
\label{Overtaking scenario}In the first scenario, an autonomous vehicle seeks to initiate a lane change maneuver, by overtaking the obstacle vehicle in the designated lane, while grappling with its level of uncertainty.
\subsubsection{Obstacle-vehicle Model}
\begin{figure*}
\centering
\begin{subfigure}{.19\textwidth}
  \centering  \includegraphics[width=0.65\linewidth]{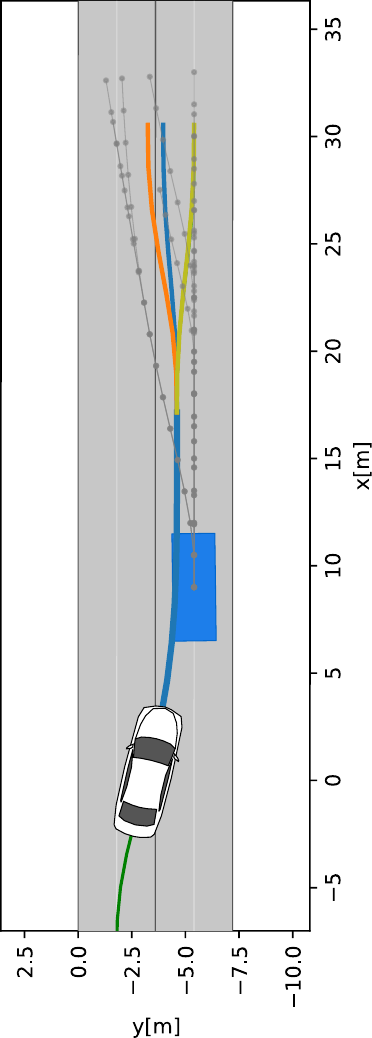}
  \caption{At $t=0.7$ s}
  \label{fig:sub1}
\end{subfigure}%
\begin{subfigure}{.19\textwidth}
  \centering
\includegraphics[width=0.65\linewidth]{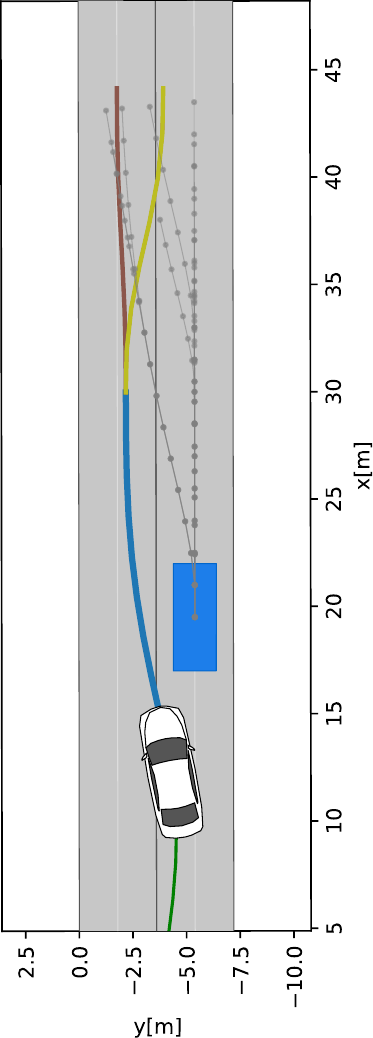}
 \caption{At $t=1.3$ s}
  \label{fig:sub2}
\end{subfigure}
\begin{subfigure}{0.19\textwidth}
  \centering
\includegraphics[width=0.65\linewidth]{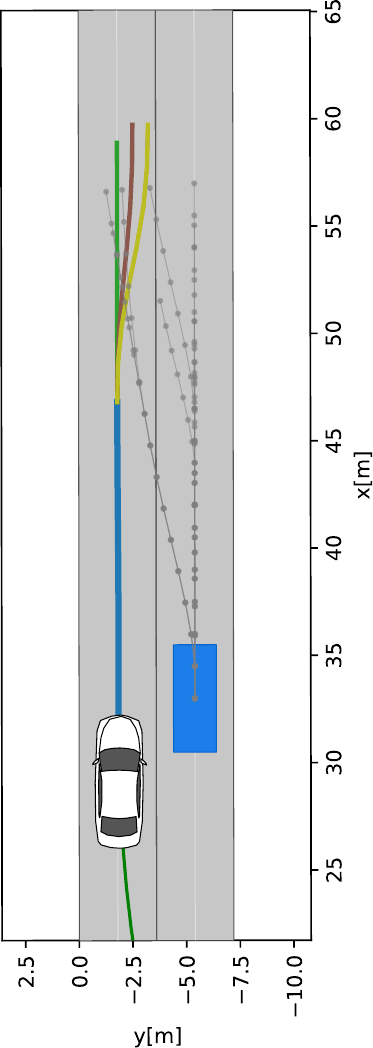}  \caption{At $t=2.2$ }
  \label{fig:sub2}
  %\hspace{-3.0cm}
\end{subfigure}
\begin{subfigure}{.19\textwidth}
  \centering
\includegraphics[width=0.65\linewidth]{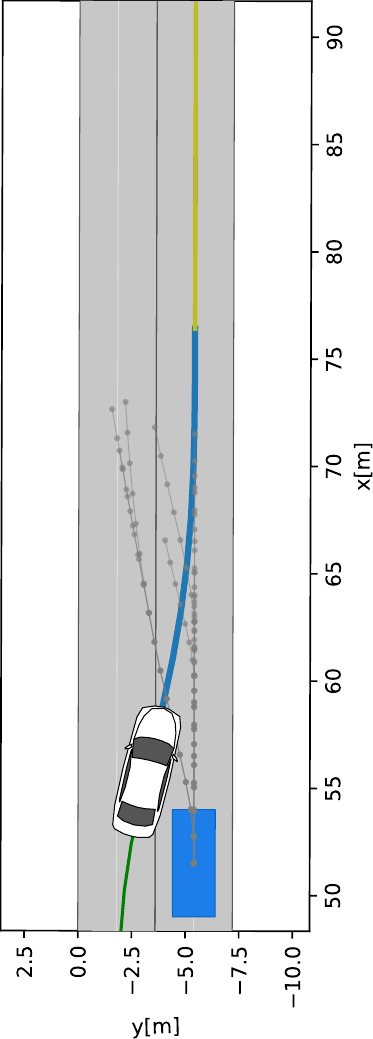}
  \caption{At $t=3.5$ s}
  \label{fig:sub2}
\end{subfigure}
\begin{subfigure}{.19\textwidth}
  \centering
\includegraphics[width=0.65\linewidth]{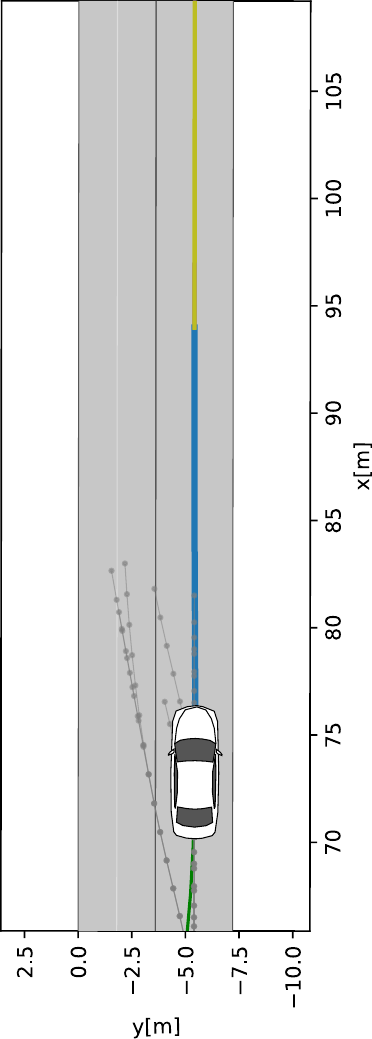}
  \caption{At $t=4.3$ s}
  \label{fig:sub2}
\end{subfigure}
\caption{Snapshots from the overtake simulated environment in CommonRoad at different time instants. The obstacle vehicle is represented by a blue rectangle. The driven trajectory by the ego-vehicle is depicted in green. The short-term planned trajectory is illustrated in blue, whereas contingent plans are depicted in distinct colors. The predicted scenario tree of the obstacle vehicle is represented in gray.}
\label{fig: scenario 1}
\end{figure*}

As a benchmark, we compare our approach to the branch-MPC introduced in \cite{Chen}. For this purpose, the same prediction model is leveraged in which it is assumed that the obstacle vehicle has three different policies that it can execute $\Lambda = \{\texttt{maintain speed}, \texttt{slow down}, \texttt{lane change}\}$ where the direction of the lane change is towards the left lane. The output of the prediction model is represented as a scenario tree, as depicted in Fig. \ref{fig: scenario 1}, in which it is assumed that the obstacle-vehicle can change its policy the next time step or after 8 steps along the horizon. In the meantime, the obstacle vehicle maintains its policy. The obstacle vehicle trajectories are constructed by forward propagating its dynamics with respect to the selected policy where the vehicle dynamics are modeled using the kinematic bicycle model \cite{car_model}. The probability of executing each policy is calculated by introducing a collision avoidance measure, $\xi(\tau, \lambda_i \in \Lambda)$, that determines the collision probability that the obstacle vehicle has, under each policy, with respect to the ego-vehicle's planned trajectory. The policy $\lambda_i$ with the least collision probability will have a higher probability of being executed by the obstacle vehicle where each probability is defined by a softmax function as described in \cite{Chen}. It is important to emphasize that although the policy of the obstacle vehicle is influenced by the planned trajectory of the ego-vehicle, the responsibility of preventing collisions rests solely upon the ego-vehicle.

\begin{table*} [ht]
\caption{Statistical results over 30 experiments. The results are reported as ``average (standard deviation).''}
\centering
\scalebox{1.0}{
\begin{tabular}{ ||c|c|c|c|c|c|c||} 
\hline
 \textbf{Planning Approach}  & \textbf{Task Duration [s]}  & \textbf{Velocity [m/s]}& \textbf{Min. Distance [m]}&
 \textbf{CR [\%]} &
 \textbf{Max. Risk} & \textbf{\textcolor{black}{Jerk [m/$\text{s}^3$]}} \\
\hline \hline
    Our method   & 3.84 (0.516) & 17.14 (0.42) &  0.61 (0.14) & 0.00 \% & 0.0491 & \textcolor{black}{2.686 (4.28)}\\
    \hline
     FISS+ \cite{FISS+}  & - & 14.87 (0.51) &  1.20 (0.00) & 0.00 \%  & 0.0393 & 2.417 (5.08)\\
     \hline
Branch-MPC \cite{Chen} & 4.68 (0.098) &  15.25 (0.45)   &  0.82 (0.04) & 0.00 \% & 0.0475 & \textcolor{black}{4.494 (5.58)}\\
\hline 
Non-cont. MLE \cite{Xu} & 3.71 (0.324) &  17.62 (0.51)   &  0.48 (0.12) & 2.74 \% & 0.1081 & \textcolor{black}{5.278 (6.13)} \\
\hline
\textcolor{black}{Contingent w/o belief updater \cite{Zhan}}& \textcolor{black}{4.46 (0.308)} &  \textcolor{black}{16.68 (0.78)}   &  \textcolor{black}{0.84 (0.08)} & \textcolor{black}{0.00 \% }& \textcolor{black}{0.0462} & \textcolor{black}{2.693 (4.63)} \\

\hline
\end{tabular}}
\label{tab:evaluation}
\end{table*}
\subsubsection{Environment Setup} To guarantee a fair comparison, for all approaches, we set the maximum speed, and maximum allowed acceleration to the same value which are 20 m$/$s, 4 $\text{m}/\text{s}^2$ respectively. The upper bound of the induced risk is assigned to $\delta  = 5\%$, for the baselines and the proposed approach, which was found to provide a good balance between safety and efficiency. A horizon of $N = 16$ steps is defined, with a discretization step of 0.2 s, resulting in a time horizon of 3.2 s, and a branching time $t_b = 1.2$ s is defined. \subsubsection{Qualitative Results} 
\begin{figure}[h]
\centering
\includegraphics[width=0.86\linewidth]{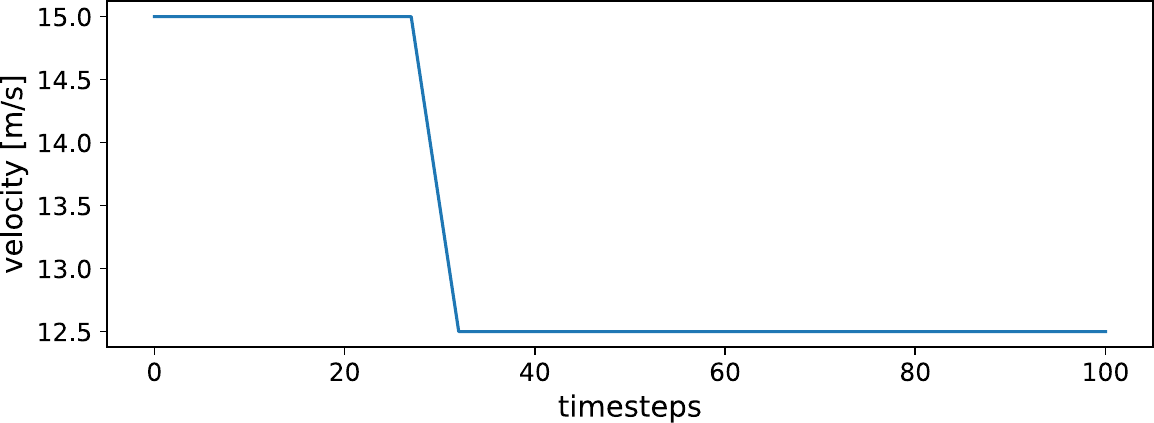}
\caption{The evolution of the obstacle-vehicle's velocity with time}
\label{obst_vel}
\end{figure}

\begin{figure}[h]
\centering
\includegraphics[width=0.86\linewidth]{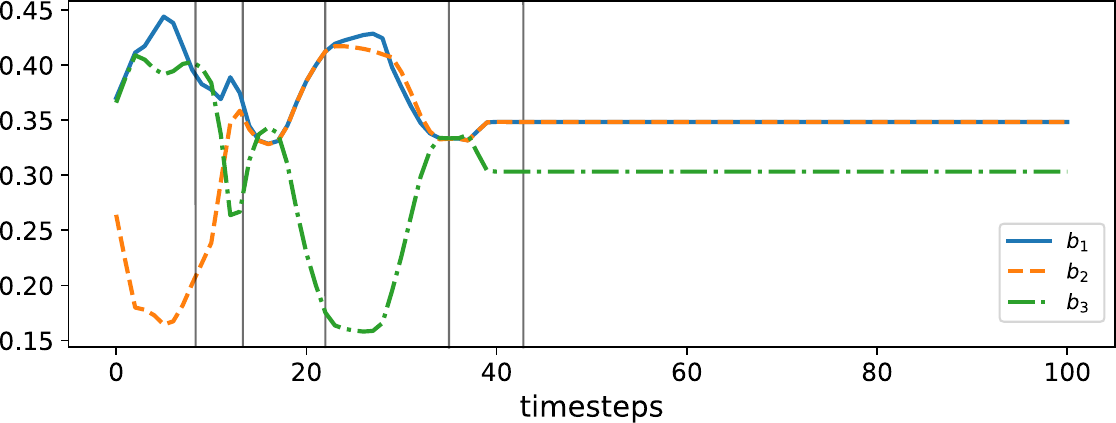}
\caption{Belief evolution over obstacle vehicle's policies along the timesteps. $b_1, b_2, b_3$ represent maintain fixed speed, slow down, and lane-change policies respectively. The gray vertical lines denote the time instants at which the snapshots in Fig. \ref{fig: scenario 1} are taken.}
\label{belief_propagation}
\end{figure}

Fig. \ref{belief_propagation} shows the evolution of the ego-vehicle's belief over the obstacle-vehicle policies over time. At the beginning, the ego-vehicle reveals its lane-change intention by swerving into the obstacle's vehicle lane. Due to the reactive behavior of the obstacle-vehicle, the probability of it executing a lane-change in the ego-vehicle's lane drops. However, the ego-vehicle could not complete the lane change since no valid trajectory is obtained that does not violate the safety constraints, and thus returns to its original lane. Since the lane-change policy of the obstacle-vehicle, $b_3$, becomes relatively low, the ego-vehicle then initiates another attempt to overtake the obstacle vehicle which probes it to decelerate allowing the ego-vehicle to complete the lane change. To visualize how the obstacle-vehicle's policy changes with the timesteps, Fig. \ref{obst_vel} shows the evolution of the obstacle-vehicle's velocity with the timesteps. As depicted, the ego-vehicle exhibits \texttt{maintain speed} policy till 2.7s. It then switches to \texttt{slow down} policy from 2.7s to 3.2s. This corresponds to the moment at which the ego-vehicle initiates its lane-change maneuver. It then switches back to the \texttt{maintain speed} policy.

On the other hand, it was observed that the branch-MPC \cite{Chen} brakes strangely while executing the lane-change maneuver relying on the obstacle-vehicle to yield to the ego-vehicle \footnote{The reader can refer to the video supplement, at 02:01, to observe such behavior.}. However, such a maneuver is risky since it may result in collisions if the obstacle-vehicle does not react on time.
\subsubsection{Quantitative Results} In each scenario, the initial state of the ego-vehicle remains fixed, while the obstacle vehicle's initial state is systematically altered across 30 distinct positions. These positions are selected from a uniform grid surrounding the nominal starting conditions. As efficiency metrics, average speed, and duration to complete the overtaking maneuver are calculated. Moreover, the average minimum distance between the ego-vehicle and the obstacle vehicle is recorded as a measure of conservatism. The quantitative results are summarized in Tab. \ref{tab:evaluation} where the \textit{non-contingent robust baseline} \cite{FISS} refers to the case in which a single plan is optimized along the entire horizon that accounts for all obstacle-vehicle policies. As shown in Tab. \ref{tab:evaluation}, our approach can complete the overtake maneuver in less duration and with a higher average speed, while providing the same safety guarantees as the branch-MPC. \textcolor{black}{This could be attributed to the fact that although the branch-MPC plans a distinct trajectory for each branch in the scenario, the optimization problem minimizes the expectation over all branches. This causes the ego-vehicle to overreact to branches with low probabilities resulting in a more conservative plan}. The non-contingent robust baseline, on the other hand, fails to complete the maneuver. Since a single trajectory is optimized that avoids all obstacle predictions, it could not find a safe trajectory to execute the lane change. As expected, the \textcolor{black}{MLE} baseline is the least-conservative approach, among the ones in comparison, since it only considers the most probable mode. This, however, results in collisions in some of the scenarios due to its \textit{over-confidence} in relying solely on the highest probable mode of the prediction model. \textcolor{black}{Our approach mitigates the limitation of the MLE method by inferring a posterior distribution over the obstacle-vehicle's intent allowing the ego-vehicle to account for uncertainty and generate safer yet efficient plans. Finally, the baseline that uses the same contingency planner as ours but lacks a belief updater, can complete the lane change maneuver safely in all experiments. Nonetheless, the absence of a belief updater prolongs the time it takes for the predictive model to assign a diminished probability to the obstacle vehicle's lane change maneuver. Consequently, this leads to a more conservative planning approach compared to the contingency planner equipped with a belief updater.} It can also be seen that all approaches respect the maximum risk threshold, $\delta = 0.05$, except the non-contingent MLE approach.
\begin{rmk}
    \textit{It is worth mentioning that, for the robust baseline, the belief updater is also utilized to weigh the evaluated collision probability of the planned trajectory with each possible mode, which, in turn, affects the calculated risk. Thus, the baseline re-plans every cycle with the newly observed obstacles' states as well.}
\end{rmk}
\begin{figure*}
\centering
\begin{subfigure}{.33\textwidth}
  \centering
  %\hspace{-1.0cm}
  \includegraphics[width=0.86\linewidth]{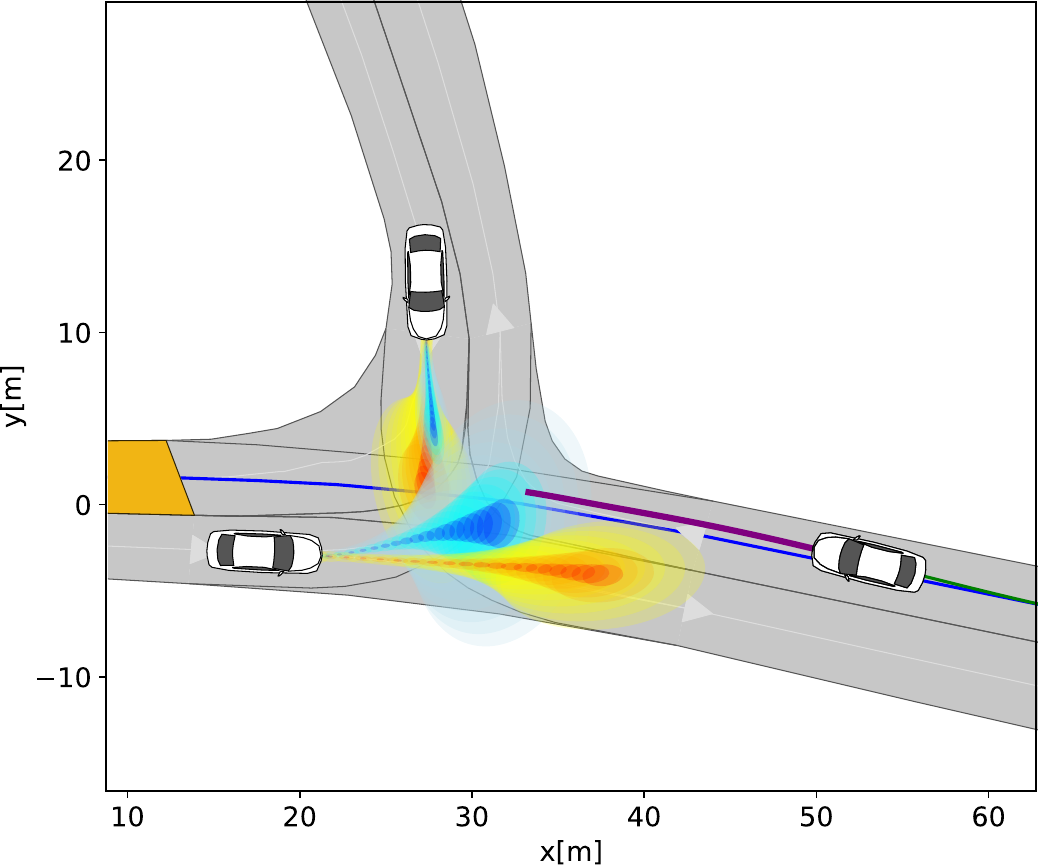}
  \caption{At $t=5.0$ s}
  \label{fig:sub1}
\end{subfigure}%
\begin{subfigure}{.33\textwidth}
  \centering
  \includegraphics[width=0.86\linewidth]{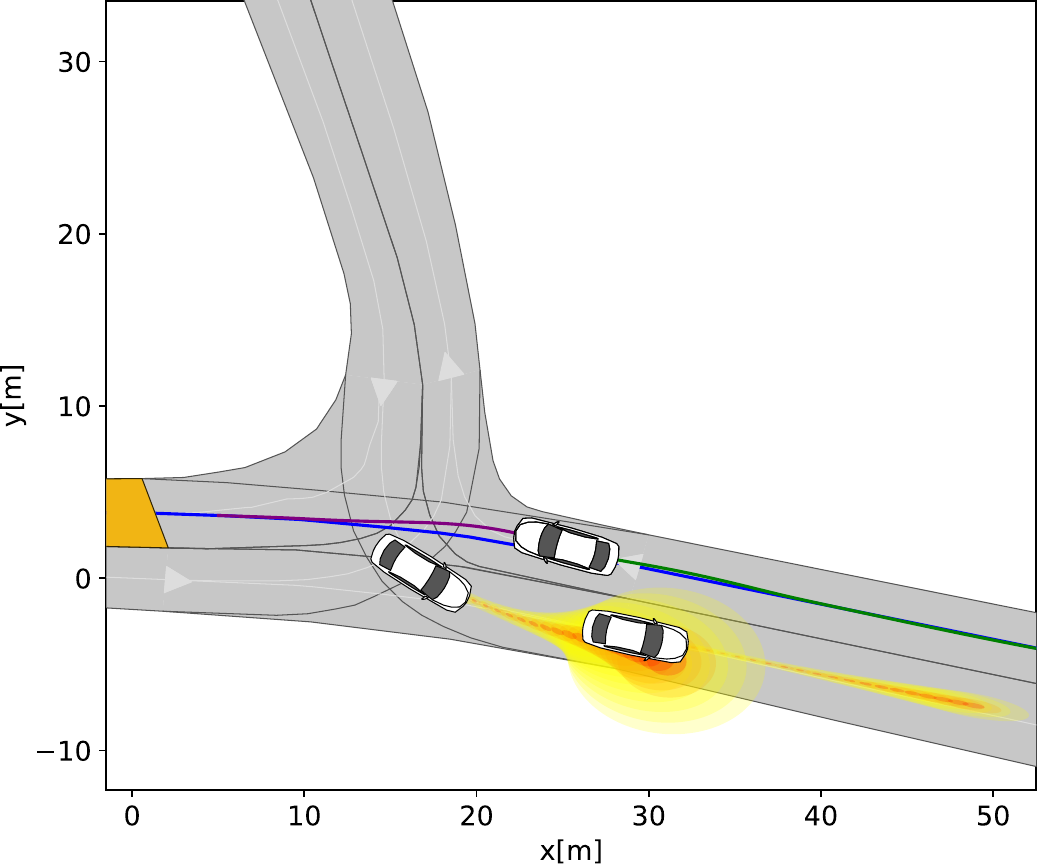}
  \caption{At $t=8.0$ s}
  \label{fig:sub2}
\end{subfigure}
\begin{subfigure}{.33\textwidth}
  \centering
\includegraphics[width=0.84\linewidth]{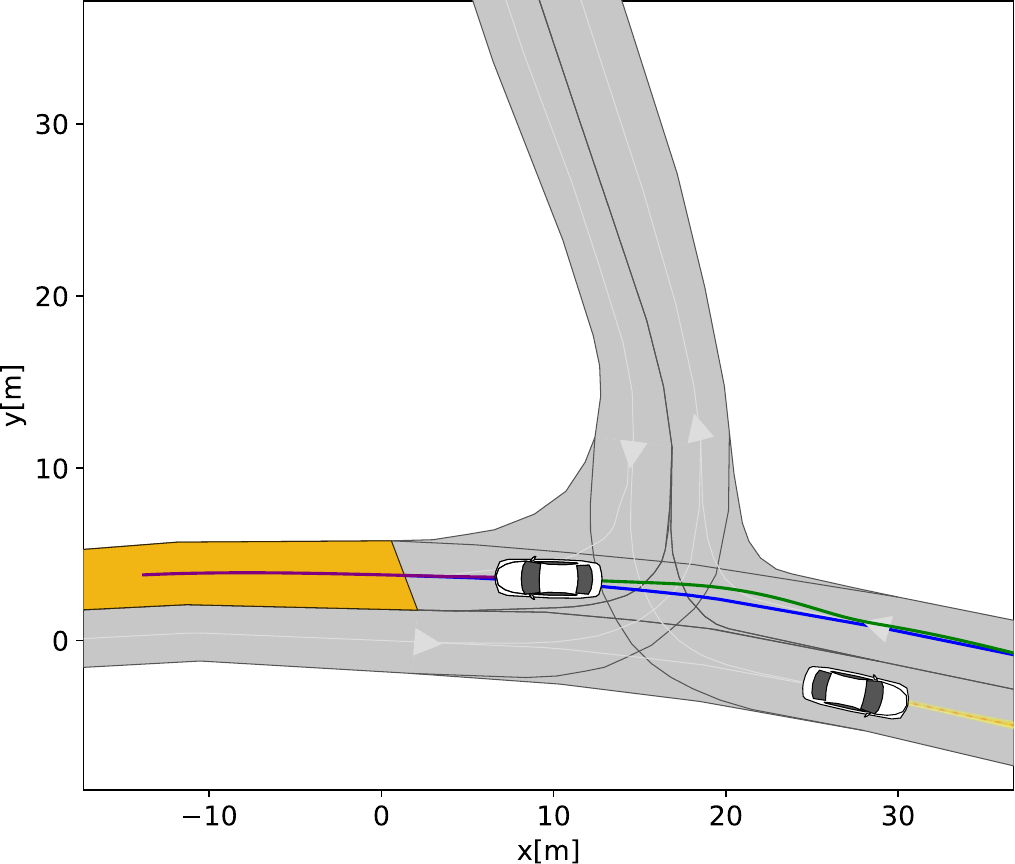}
  \caption{At $t=10.0$ s}
  \label{fig:sub2}
  %\hspace{-3.0cm}
\end{subfigure}
\caption{Snapshots from the T-junction simulated environment in CommonRoad at different time instants. The ego-vehicle's plan is to drive from right to left while avoiding collisions with the other two obstacle vehicles. In Fig 9(a), the prediction of both modes of both obstacles is represented by colored ellipsoids where the blue ellipsoids depict the \texttt{left turn} and \texttt{yield} policies for the leftward and upward vehicles respectively whereas the yellow ellipsoids illustrate the \texttt{lane keep} and \texttt{left turn} policies. The driven trajectory by the ego-vehicle is depicted in green. The short-term planned trajectory is illustrated in purple.}
\label{fig:T-junction}
\end{figure*}
\subsection{Scenario 2. Urban T-junction:}
\label{T-junction scenario}
In this scenario, the ego vehicle is approaching a T-junction, \textit{with no traffic rules}, in which its mission is to follow its designated lane while being uncertain about the intentions of the other vehicles as shown in Fig. \ref{fig:T-junction}. Here, we consider the case of a multi-vehicle traffic scenario in which two obstacle vehicles approach the T-junction where $\Lambda_1 = \{\texttt{lane keep}, \texttt{left turn}\}$, and $\Lambda_2 = \{\texttt{left turn}, \texttt{yield}\}$. In this case, it is not sufficient to consider the belief of a single agent's intention as we did in the previous scenario, however, instead, we need to get a belief about how the traffic scene will evolve by considering all permutations the traffic participants can have as stated in \eqref{permutation}.
\subsubsection{Environment Setup} The states of both obstacles are randomly initialized and their corresponding policies are randomly assigned from the set of potential policies. The initial velocities of the obstacles are selected in such a way that they arrive at the intersection before the ego-vehicle, forcing the ego-vehicle to react and avoid collisions actively. 
\begin{figure}[h!]
  \centering
  \begin{subfigure}{0.8\columnwidth}
    \includegraphics[width=\linewidth]{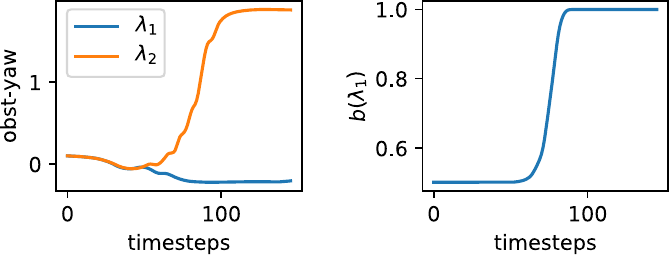}
    \caption{}
    \label{belief-obs1}
  \end{subfigure}
  
  \begin{subfigure}{0.8\columnwidth}
    \includegraphics[width=\linewidth]{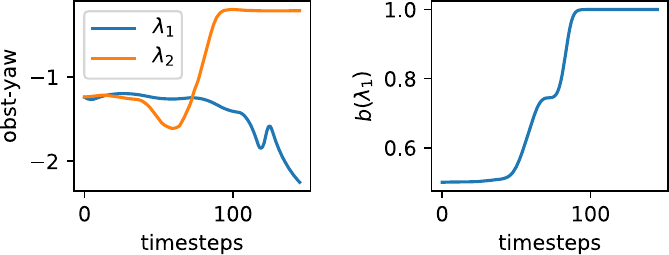}
    \caption{}
    \label{belief-obs2}
  \end{subfigure}
  \caption{The evolution of the belief of both obstacles based on their observed orientations according to \eqref{belief}. Top: the belief evolution for obstacle 1 where $\lambda_1$ represents the lane keep policy and $\lambda_2$ represents the left turn policy. Bottom: the belief evolution for obstacle 2 where $\lambda_1$ represents the yield policy whereas $\lambda_2$ represents the left turn policy. Note that this represents one of the four permutations that the traffic scene can evolve to.}
  \label{fig:two-subplots}
\end{figure}

\begin{figure}[h]
\centering
\includegraphics[width=0.86\linewidth]{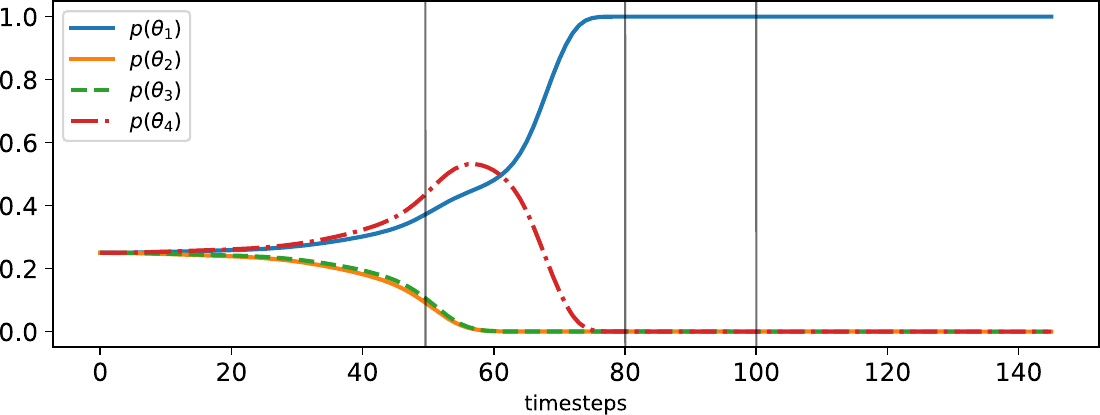}
\caption{The evolution of the belief over permutations in the T-junction scenario where $\theta_1$ corresponds to \texttt{lane keep} and \texttt{left turn} for the leftward and upward vehicles respectively. The gray vertical lines denote the time instants at which the snapshots in Fig. \ref{fig:T-junction} are taken.}
\label{permutations evolution}
\end{figure}
\noindent In this scenario, a horizon of $N = 25$ steps is defined, with a discretization step of 0.2 s, resulting in a time horizon of 5.0 s, and a branching time $t_b = 2.4$ s is defined. As a benchmark comparison, we evaluated the same task using a robust planner that optimizes a single plan that considers all the modes that the other agents could have, and a greedy baseline that only considers the most probable predicted mode of each obstacle. All optimization parameters for both methods are set to be identical to guarantee a fair comparison. At the start of the simulation, all permutations, $\theta_i \in \Theta$, are initialized with equal likelihoods, $\theta_i = 0.25$. The evolution of the belief over both obstacle modes is illustrated in Fig. \ref{fig:two-subplots}. In Fig. \ref{fig:two-subplots}, we present the dynamic evolution of the ego-vehicle's belief regarding various intentions of surrounding obstacles over time. Specifically, Fig. \ref{belief-obs1} illustrates this evolution for a leftward-bound vehicle encountering a T-junction, where the obstacle vehicle faces the choice between continuing straight or executing a left turn. Similarly, Fig. \ref{belief-obs2} portrays the belief dynamics for an upward-bound vehicle confronted with the options of turning left or yielding to the ego-vehicle. In Fig. \ref{belief-obs1}, we observe the ego-vehicle's initial struggle with uncertainty regarding the leftward vehicle's intentions, reflected in an equal belief distribution ($b(\lambda_1)=b(\lambda_2) = 0.5$) as its state aligns with the mean of both distributions. However, as the obstacle vehicle's state gradually deviates from this equilibrium, the belief over the left-turn maneuver diminishes, leading to a corresponding increase in belief regarding the alternative mode ($b(\lambda_1)$). This nuanced adjustment allows the ego-vehicle to attenuate its emphasis on the left-turn possibility, thereby facilitating an accelerated trajectory within the T-junction. Analogous dynamics are observed for the second obstacle vehicle, as depicted in Fig. \ref{belief-obs2}. Subsequently, the updated beliefs per mode, $\lambda \in \Lambda$, are utilized to update the probabilities over the different permutations, $\theta \in \Theta$, the traffic scene can evolve to as shown in Fig. \ref{permutations evolution}. As illustrated in Fig. \ref{permutations evolution}, the evolution of permutations depicts a gradual decrease in the belief regarding $\theta_2$ and $\theta_3$ over successive iterations, ultimately diminishing after approximately 58 time steps where their influence on the ego-vehicle's plan is disregarded. Furthermore, by the 76th time-step, the ego-vehicle attains a high level of certainty that $\theta_1$ is the accurate hypothesis adopted by the obstacles. As a result, only this prediction mode significantly impacts the planner's decision-making process.
\begin{table*}[ht]
\caption{Statistical results over 100 experiments for a T-junction scenario. The comparison is done with respect to the traveled distance by the ego-vehicle, ego-vehicle velocity, minimum distance to the obstacles, and the maximum deployed acceleration. The results are reported as ``average (standard deviation)''. The branching time is set to $t_b = 2.4$ s.}
\centering
\scalebox{1.0}{
\begin{tabular}{ ||c|c|c|c|c|c|c||}
 \hline
 \textbf{Approach} & \textbf{Progress [m]} & \textbf{Velocity [m/s]} &  \textbf{Min. Distance. [m]} & \textbf{Max. Acc. [$\text{m}/\text{s}^2$]} & \textbf{CR[\%]} & \textbf{\textcolor{black}{Jerk [m/$\text{s}^3$]}} \\
 \hline \hline
 Our method   & 91.324 (0.169) &  7.017 (0.133) & 1.58 (0.06) & 2.931 (0.436) & 0.00 \% & \textcolor{black}{0.841 (0.71)}\\
 \hline
  FISS+ \cite{FISS+}   & 83.216 (0.413) &  6.126 (0.264) & 2.08 (0.08) & 1.243 (0.617) & 0.00 \% & 0.716 (0.42)\\
\hline
  Non-cont. MLE \cite{Xu}   & 96.581 (0.648) &  7.443 (0.338) & 1.06 (0.18) & 1.424 (0.556) & 3.94 \% & \textcolor{black}{0.688 (0.29)}\\
  \hline
\textcolor{black}{Contingent w/o belief updater \cite{Zhan}}& \textcolor{black}{85.649 (0.372)} &  \textcolor{black}{6.581 (0.162)}   &  \textcolor{black}{1.94 (0.11)} & \textcolor{black}{2.983 (0.632) }& \textcolor{black}{0.00 \%} & \textcolor{black}{1.438 (0.96)} \\
\hline

\end{tabular}}
\label{tab:T-junction}
\end{table*}
\subsubsection{Prediction model} In this scenario, we use a synthesized prediction model that incorporates the multi-modality in the obstacle-vehicle's intentions. Specifically, our multi-modal prediction model works as follows:
\begin{enumerate}[(i)]
    \item By identifying target lanes in the T-junction, we extrapolate the intended trajectories of surrounding vehicles.
    \item By employing a motion planner for each vehicle, we generate ground truth trajectories towards these lanes, resulting in multi-modal trajectories per vehicle.
    \item Our multi-modal prediction strategy involves:
    \begin{itemize}
        \item Utilizing a uni-modal prediction model trained on extensive CommonRoad datasets \cite{watchandlearn} to generate Gaussian trajectory distributions for each potential mode, corresponding to the trajectories from Step (ii).
        \item Amalgamating these distributions into a Gaussian Mixture Model (GMM), weighted by their likelihoods, inspired by prior works such as \cite{Bajcsy2, Bansal}.
    \end{itemize}
\item Mode weights, determining the likelihood of each trajectory mode, are computed based on collision avoidance metrics. These metrics, quantifying collision probabilities between obstacle vehicle trajectories and the ego-vehicle's planned trajectory, dynamically adjust mode weights using a softmax function inspired by works such as \cite{Chen}. 
\end{enumerate}
\begin{rmk}
    \textit{We emphasize that our approach is agnostic to the prediction model employed. Any prediction model capable of providing Gaussian distributions over the predicted modes can be utilized.}
\end{rmk}
\subsubsection{Quantitative Results} 
\begin{figure}[h!]
\centering
\includegraphics[width= \linewidth]{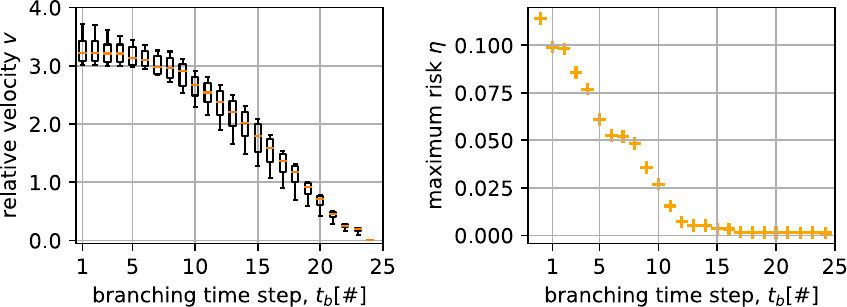}
\caption{Left: the performance gap regarding the relative velocity for different branching times, right: the maximum risk, $\eta$, recorded in all simulations for different branching times.}
\label{branching time analysis}
\end{figure}

The quantitative results are reported in Tab. \ref{tab:T-junction}. The significant enhancement in the ego-vehicle's performance is attributed to its ability to delay the braking decision, thanks to multiple contingent plans, as long as it is capable of safely braking later when it gets more certainty about other obstacles' intents to react to any possible outcome. This, as expected, comes at the expense of stopping closer to the obstacle, and braking more aggressively in situations in which the ego-vehicle has to yield to the obstacles. Despite this delayed decision-making, the maximum risk encountered by the ego-vehicle, in the contingency planning case, is still significantly below the defined upper-bound in the chance constraints, as depicted in Fig. \ref{branching time analysis}, showing that performance improvement is attained without compromising safety. As in the previous scenario, the MLE approach has the best performance in terms of average velocity and progress along the driving route. This is, however, achieved at the expense of resulting in collisions making it not safe to be deployed. \textcolor{black}{Similar to the overtaking scenario, the contingency planning without a belief updater has a less efficient performance compared to our proposed approach showing that the belief updater improves the planner's performance without compromising safety.}

Here it is worth pointing out that the proposed algorithm is implemented in Python to interface with CommonRoad. The average computational time over the experiments is 174.98 ms. Tab. \ref{tab:time analysis} shows how the computational time $t_c$ scales with the number of agents.  We emphasize that since the sampling-based approach is parallelizable, the computational time can be further improved by evaluating the constraints of the sampled trajectories through parallelizable computations.
\begin{table} [h!]
% Set the color of the table caption label
\centering
\scalebox{1.0}{
\begin{tabular}{ ||c|c|c|c||} 
\hline
{\textbf{Number of agents}}  & {\textbf{1}} & {\textbf{2}} & {\textbf{3}} \\
\hline \hline
   {$t_c$ (ms)}  & {134.53} &  {186.12} & {204.31} \\
      {$\sigma_{t_c}$ (ms)}  & {4.7} &  {6.9} & {11.8} \\
    \hline
\end{tabular}}
\caption{Average computational time $t_c$ and standard deviation $\sigma_{t_c}$ for increasing number of agents.}
\label{tab:time analysis}
\end{table}

So far, we considered a certain value that we assign to the branching time $t_b$. In the case of \textit{open-loop} planning, as in \cite{Hardy, Bajcsy}, the branching time is not an independent design parameter, and it has to be estimated correctly, otherwise, the ego-vehicle will branch to an over-confident contingent plan by $t_b$ which can result in a collision. In our proposed approach, however, thanks to planning in a \textit{closed-loop} with a \textit{belief updater}, the branching time does not need to be estimated exactly. However, low branching times may lead the ego-vehicle to inevitable states from which it could not recover in case of certain obstacles' permutations, due to the limited dynamics capabilities. Thus, restrictions still apply when it comes to assigning a branching time which we discuss in the following section.
\subsubsection{Effect of branching time on the plan}
\label{sec:branching time}
In this section, an analysis of how the branching time affects contingency planning is conducted. For this purpose, we run experiments for all values of branching time $t_b \in [\Delta t, T]$, where $\Delta t$ is the discretization step that we set to 0.2 s. For each branching time, we run 100 simulated experiments in which the obstacles' intents and their initial states are randomly initialized. We analyze the effect of the branching time on the relative average velocity the ego-vehicle exhibits with respect to the baseline, $t_b = T$. Moreover, the maximum risk among all experiments for each branching time is recorded. The results are reported in Fig. \ref{branching time analysis}. As shown, for larger values of the branching time, the performance gap between both methods is small since most of the plan is constituted by the shared plan and thus the effect of the belief updater in the cost function is not pronounced.
Indeed, when $t_b=T$, the disparity in performance disappears, as our proposed approach aligns with the baseline method under such conditions. For earlier branching times, however, the performance gap becomes more pronounced since the future information gain beyond the branching time is well exploited in the cost function because of the additional degrees of freedom introduced by the contingent plans. By inspecting the maximum risk plot depicted in Fig. \ref{branching time analysis}, it can be observed that the maximum risk, $\eta$, increases as the branching time becomes shorter. This can be attributed to the over-confidence in the planned trajectory after the branching time causing the ego-vehicle to take more risky maneuvers. For sufficiently short branching times, $t_b \leq 2.0$ s in this example, the ego-vehicle could not find a feasible trajectory that does not violate the maximum risk, $\delta = 0.05$, in the chance constraint, and subsequently, we execute the planned with the least risk that is dynamically feasible as we indicated earlier in Remark \ref{invalid}. 
\begin{figure*}
\centering
\begin{subfigure}{.33\textwidth}
  \centering
  %\hspace{-1.0cm}
  \includegraphics[width=0.86\linewidth]{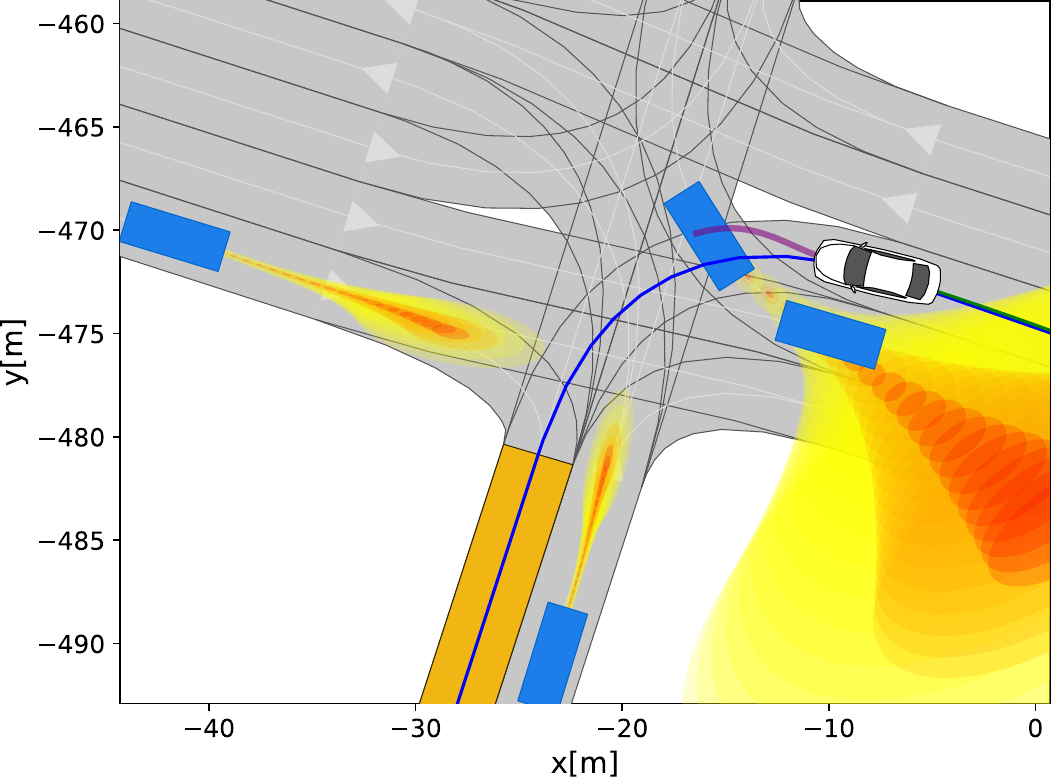}
  \caption{At $t=2.8$ s}
  \label{fig:sub1}
\end{subfigure}%
\begin{subfigure}{.33\textwidth}
  \centering
  \includegraphics[width=0.86\linewidth]{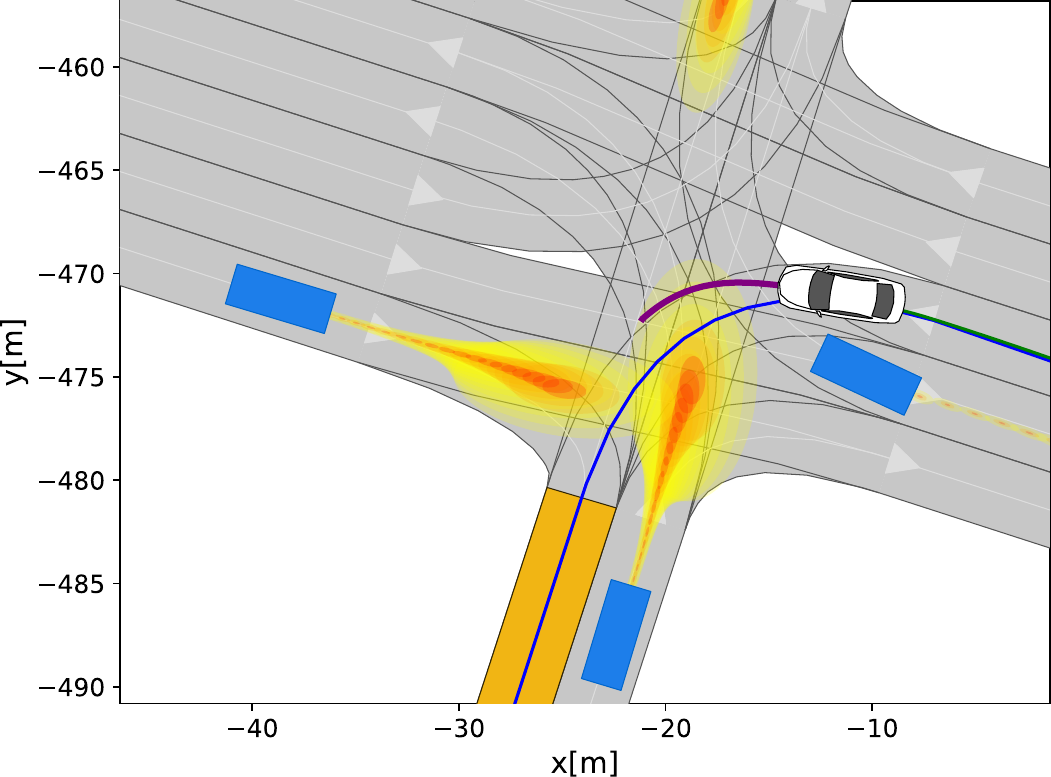}
  \caption{At $t=3.6$ s}
  \label{fig:sub2}
\end{subfigure}
\begin{subfigure}{.33\textwidth}
  \centering
\includegraphics[width=0.86\linewidth]{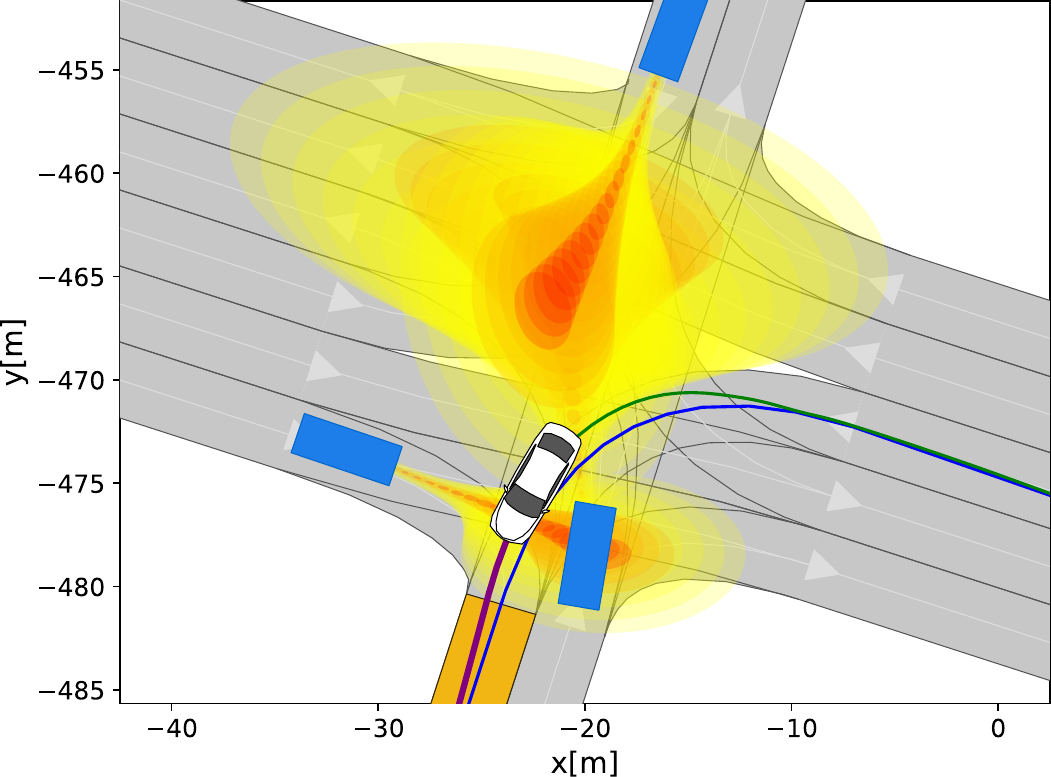}
  \caption{At $t=5.2$ s}
  \label{fig:sub2}
  %\hspace{-3.0cm}
\end{subfigure}
\caption{Snapshots from the intersection simulated environment in CommonRoad at different time instants. The ego-vehicle's plan is to take a left turn while avoiding collisions with the other two obstacle vehicles. The ego-vehicle is depicted as a car icon whereas the obstacle vehicles are represented by blue rectangles. The prediction of the most probable mode of the obstacle vehicles is represented by colored ellipsoids. The driven trajectory by the ego-vehicle is depicted in green. The short-term planned trajectory is illustrated in purple.}
\label{fig:simulation}
\end{figure*}
This concludes that the branching time in contingency planning is related to the maximum risk the ego-vehicle perceives. More analysis regarding the estimation of the branching time $t_b$ based on the ego-vehicle's dynamics capabilities is left for future work.
\subsubsection{\textcolor{black}{Branching time estimation}}\textcolor{black}{In this section, we examine the effect of updating the branching time online, based on the updated belief the ego-vehicle maintains over different prediction modes, compared to fixing the branching time $t_b$ to a certain value. To do so, three different baselines are considered.\\
\textbf{Baseline 5: ``Oracle'' branching time.} This estimator reconstructs the actual branching time by initially simulation the planning problem using a nominal branching time. Subsequently, it extracts the moment of certainty from the retrospective evolution of beliefs. It is important to note that the oracle relies on access to the true human intent and, as such, is not implementable in real-world scenarios. Nevertheless, we incorporate this variant to illustrate the potential performance attainable with the true branching time.\\
\textbf{Baseline 6: Branching time heuristics adopted from \cite{Peters}.} This heuristic considers the entropy of the belief that the ego-vehicle maintains over each hypothesis, $\theta \in \Theta$, as an indication of how the observed obstacles' states are distinct, 
\begin{equation}
    \mathcal{H}(p(\theta_i))=-\sum_{\theta_i \in \Theta} p(\theta_i) \log_{|\Theta| }(p(\theta_i))
\end{equation}
To estimate the branching time, the predicted trajectories of each obstacle along each hypothesis from the previous time-step, $\boldsymbol{\delta}_{\theta, t-1}^o$ are considered as hypothetical observations that can be inferred from their prediction model to estimate the associate belief according to \eqref{belief}. Another operator, $B(\boldsymbol{\delta}_{\theta, t-1}^o, \theta, k)$, is introduced that takes as input the first $k$ steps from the hypothetical observation, $\boldsymbol{\delta}_{\theta, t-1}^o$, and returns the updated belief. Accordingly, the branching time is estimated as 
\begin{equation}
\begin{aligned}
    t_b = \max_{\theta \in \Theta} & \min_{ k \in \{2, ..., T\}} \quad \omega.k \\
    & \text{s.t.} \quad \quad \quad \mathcal{H}[B(\boldsymbol{\delta}_{\theta, t-1}^o, \theta, k)] \leq \epsilon
\end{aligned}
\end{equation}
where $\omega$ indicates the discretization step. This heuristic estimates the branching time as the first time at which all predicted beliefs reach a certain threshold $\epsilon$, assuming that the obstacles behave rationally with respect to their prediction models.\\
\textbf{Baseline 7: Branching time heuristic adopted from \cite{Marc}.} This heuristic chooses the branching time as the maximum time such that any two future scenarios starting at the current time only diverge by a maximum distance,
\begin{equation}
\begin{aligned}
    t_b = \max_{k\in {2,...,T}} \quad & \omega.k \\
    & \text{s.t.} \quad \mathcal{M}(\theta, k) \leq \epsilon, \quad \forall \theta \in \Theta
\end{aligned}
\end{equation}
where $\mathcal{M} $ represents the divergence measure. This heuristic, however, entails at least double the computational time since the divergence measure is invoked on the ego-vehicle trajectories.\\
For all baselines, the branching time is updated at every time step and a Monte Carlo study is conducted to analyze the performance of updating the branching time at every time step compared to fixing the branching time to a certain value, $t_b =2.4$, that we used in the evaluations in Sections \ref{Overtaking scenario} and \ref{T-junction scenario}.} 
\begin{table} [h!]
% Set the color of the table caption label
\centering
\scalebox{1.0}{
\begin{tabular}{ ||c|c|c|c||} 
\hline
 \textcolor{black}{\textbf{Approach}}  & \textcolor{black}{\textbf{Progress [m]}} & \textcolor{black}{\textbf{Velocity [m/s]}} & \textcolor{black}{\textbf{Max. Risk}} \\
\hline \hline
   \textcolor{black}{Oracle}  & \textcolor{black}{91.688 (0.178)} &  \textcolor{black}{7.045 (0.132)} & \textcolor{black}{0.0464} \\
    \textcolor{black}{$t_b = 2.4$ (Ours)}  & \textcolor{black}{91.324 (0.169)} & \textcolor{black}{7.017 (0.133)} &  \textcolor{black}{0.0461}\\
 \textcolor{black}{Heuristics in \cite{Peters}} & \textcolor{black}{89.948 (0.148)} & \textcolor{black}{6.912 (0.192)}    &  \textcolor{black}{0.0426}\\
\textcolor{black}{Heuristics in \cite{Marc}} & \textcolor{black}{88.836 (0.163)} & \textcolor{black} {6.825 (0.151)} & \textcolor{black}{0.0418}  \\
\hline
\end{tabular}}
\caption{\textcolor{black}{Statistical results over 100 experiments for a T-junction scenario with different branching time estimates.}}
\label{tab:branching}
\end{table}

 \textcolor{black}{As shown in Tab. \ref{tab:branching}, the performance gap between fixing the branching time to a certain value, and using an oracle estimate is very small. This can be
attributed to the fact that since the short-term plan cost is weighted by the belief the ego vehicle maintains over the long
plans, the short-term plan tends to be biased toward the
long-term plan with the highest belief.}

\subsection{Scenario 3. Intersection}
In this scenario, the ego-vehicle is tasked with executing a left turn within an urban intersection, all while navigating interactions with multiple obstacle vehicles simultaneously.  Each of these obstacle vehicles within the intersection has the option to either yield to the ego-vehicle, thereby allowing it to complete its left turn unimpeded, or to challenge the ego-vehicle and take priority, thereby compelling the ego-vehicle to yield. Similar to the T-junction scenario, we evaluate the efficacy of our proposed approach against established baselines. Quantitative results are presented in Tab \ref{tab:intersection}. For consistency, we utilize a horizon of $N=25$ steps, with a discretization step of 0.2 s, resulting in a time horizon of 5.0 s. Additionally, a branching-time of $tb = 2.4$ s is defined. Here it should be noted that the prediction model in this scenario is similar to the one employed in the T-junction scenario.

\begin{table} [h!]
% Set the color of the table caption label
\centering
\scalebox{0.95}{
\begin{tabular}{ ||c|c|c|c||} 
\hline
 {\textbf{Approach}}  & {\textbf{Progress [m]}} & {\textbf{Velocity [m/s]}} & {\textbf{CR}} \\
\hline \hline
   {Our method}  & {40.696 (0.582)} &  {8.139 (0.147)} & {0.00 \%} \\
    {FISS+ \cite{FISS+}}  & {29.225 (0.473)} & {5.851 (0.168)} &  {0.00 \%}\\
{Non-cont. MLE \cite{Xu}} & {42.261 (0.327)} & {8.452 (0.263)}    &  {3.47 \%}\\
{Cont. w/o belief updater} & {37.693 (0.372)} & {7.531 (0.126)} & {0.00 \%}  \\
\hline
\end{tabular}}
\caption{{Statistical results over 100 experiments for the intersection scenario where CR refers to the collision rate.}}
\label{tab:intersection}
\end{table}

{The outcomes of this scenario mirror those of previous ones, demonstrating that our proposed approach enables the ego-vehicle to successfully execute the left-turn maneuver with enhanced efficiency compared to the robust baseline, where the ego-vehicle is required to yield while waiting for other vehicles to execute their maneuvers inside the intersection\footnote{{The reader can refer to the video supplement to observe the yielding behavior of the ego-vehicle using the robust baseline.}}. However, it's noteworthy that the MLE approach outperforms our method in terms of performance, albeit at the cost of a higher collision rate, as it only considers the most probable mode without accounting for potential deviations.}

\section{Conclusion}
%The conclusion goes here.
This paper introduced a novel contingency planning framework that integrates the ego-agent's beliefs regarding the potential multi-modal behaviors exhibited by surrounding agents. This belief is continuously updated based on inferred states of observed obstacles from a predictive model. The methodology involves decomposing the planning task into short-term and long-term plans, with each long-term plan being tailored to a specific obstacle policy. The resultant contingency plans contribute to the overall plan's cost by factoring in their costs along with the associated belief values derived from the belief updating process. The effectiveness of the proposed approach was evaluated in the context of two safety-critical driving scenarios. Through comprehensive closed-loop simulations, we compared our proposed planner against different baselines. We demonstrated that our approach achieves less conservative driving behavior compared to a state-of-the-art multi-policy algorithm while maintaining equivalent safety assurance. Our approach has also outperformed the traditional planner that optimizes over all possible modes provided by a prediction model. To analyze the effect of the Bayesian belief updater on contingency planning, we showed that the belief updater improves the planner's performance without compromising safety. The influence of branching time on the planner's performance was investigated, and the adaptability of the proposed approach to scenarios involving multiple vehicles was explored.

{\appendix[\textcolor{black}{Baselines Comparison}]
\textcolor{black}{To ensure a fair comparison between our proposed contingency planning approach and the branching-MPC approach proposed in \cite{Chen}, the following measures are considered. Except for the branch-MPC \cite{Chen}, all planners employ the ego-motion sampler detailed in Section \ref{sampler} in the Frenet frame with the same cost function and constraints to rank the generated samples. Nevertheless, although the branch-MPC utilizes a different planner, we modified the cost function such that it is aligned closely to the one used with the Frenet planner. The utilized cost function for the Frenet planner is given as,}
\begin{equation*}
    \textcolor{black}{J(\tau)= w_v c_v + w_d c_d + w_a c_a + w_{\dot{\delta}} c_{\dot{\delta}}}
\end{equation*}
\textcolor{black}{where $c_v, c_d$ are the costs for velocity and reference tracking, whereas $c_a, c_{\dot{\delta}}$ penalize the acceleration and steering angle rate respectively. On the other hand, for the branch-MPC, similar to the original paper, a unicycle model is adopted where the states are given by $x=[X, Y, v, \psi]^T$, and the inputs $u=[a, \dot{\delta}]$. The cost function for the branch-MPC is, accordingly, defined as}
\begin{equation*}
    \textcolor{black}{J(\tau)=(x-x_{\text{ref}})^TQ(x-x_{\text{ref}})+u^TRu}
\end{equation*}
\textcolor{black}{where $Q=\texttt{diag}(0, w_d, w_v, 0)$, and $R=\texttt{diag}(w_a, w_{\dot{\delta}})$. In this way, we ensure that the cost functions used by the Frenet and branch-MPC planners are similar. Additionally, the same kinematic constraints are applied to all planners including the curvature 
constraints and the box constraints imposed on the velocity, acceleration, and jerk. Tab. \ref{tab:parameters} summarizes the parameters utilized by the planners in the evaluations. %The source code of our planner is available in this \href{https://github.com/KhMustafa/Risk-aware-contingency-planning-with-multi-modal-predictions}{repository}.
\begin{table} [h!]
\centering
\scalebox{1.0}{
\begin{tabular}{ ||c|c|c|c||} 
\hline
{\textbf{Description}}  & {\textbf{Notation}} & {\textbf{Value}} & {\textbf{Unit}} \\
\hline \hline
{Maximum velocity}  & {$v_{\text{max}}$} & {20} &  {m/s}\\
{Minimum velocity} & {$v_{\text{min}}$} & {0}    &  {m/s}\\
{Maximum acceleration} & {$a_{\text{max}}$} &  {4} & {$\text{m}/\text{s}^2$}  \\
 {Lane width}  & {$lw$} &  {3} & {m} \\
   {Receding horizon}  & {$N$} &  \textcolor{black}{16} & \textcolor{black}{-} \\
        \textcolor{black}{Discretization step}  & \textcolor{black}{$\omega$} &  \textcolor{black}{0.2} & \textcolor{black}{-} \\
        \textcolor{black}{Branching time}  & \textcolor{black}{$t_b$} &  \textcolor{black}{1.2} & \textcolor{black}{s} \\
         \textcolor{black}{Risk threshold}  & \textcolor{black}{$\delta$} &  \textcolor{black}{5} & \textcolor{black}{\%} \\
         \textcolor{black}{Vehicle width}  & \textcolor{black}{$W$} &  \textcolor{black}{2.5} & \textcolor{black}{m} \\
            \textcolor{black}{Vehicle length}  & \textcolor{black}{$L$} &  \textcolor{black}{4} & \textcolor{black}{m} \\
  \textcolor{black}{Speed cost weight}
            & \textcolor{black}{$w_v$} &  \textcolor{black}{1} & \textcolor{black}{-} \\
              \textcolor{black}{Lat. deviation cost weight}
            & \textcolor{black}{$w_d$} &  \textcolor{black}{1} & \textcolor{black}{-} \\
            \textcolor{black}{Acceleration cost weight}
            & \textcolor{black}{$w_a$} &  \textcolor{black}{2} & \textcolor{black}{-} \\
            \textcolor{black}{Steering rate cost weight}
            & \textcolor{black}{$w_{\dot{\delta}}$} &  \textcolor{black}{1} & \textcolor{black}{-}\\ 
\hline
\end{tabular}}
\caption{\textcolor{black}{List of parameters utilized by the planners for the overtaking scenario evaluations.}}
\label{tab:parameters}
\end{table}
}

{\appendices
\section*{{Branching Time Selection}}
{As mentioned in Section \ref{sec:branching time}, the branching time is a critical parameter that affects the contingency planning efficiency. To justify our branching time selection for the simulated scenarios in Section \ref{results}, we conducted an ablation study that measures the performance gap between different branching times and an oracle that has access to the true human intent.
Fig. \ref{fig: branching time selection} illustrates the performance gap regarding the relative velocity for different branching times for both the overtake and T-junction scenarios. As shown, for both scenarios, for the branching times in the middle of the planning horizon, the performance gap compared to the oracle gets smaller. As the branching time gets smaller, a less conservative approach compared to the oracle can be achieved. This comes, however, at the expense of having higher risk as we discussed earlier in Section \ref{sec:branching time}. In contrast, with large branching times, the contingency planning becomes more conservative compared to the oracle. Based on the obtained empirical results, we can conclude that fixing the branching time to a certain value in the middle of the planning horizon can achieve close performance to the updating it based on an oracle in hindsight.}
\begin{figure}[h!]
\centering
\includegraphics[width= \linewidth]{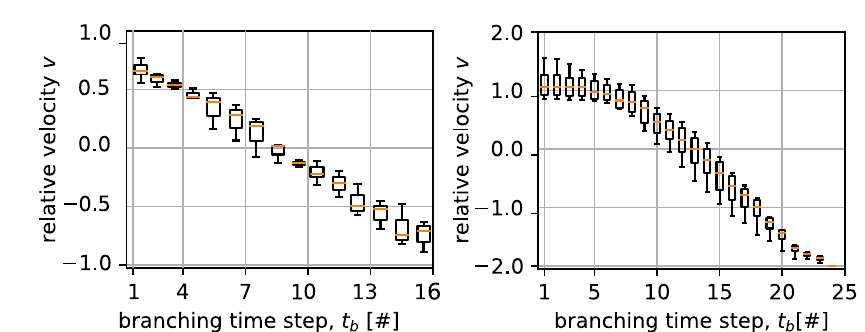}
\caption{{The performance gap regarding the relative velocity for different branching times. Left: overtake scenario, right: T-junction scenario.}}
\label{fig: branching time selection}
\end{figure}
}

\vfill

\end{document}